\documentclass{article}

\usepackage{microtype}
\usepackage{graphicx}
\usepackage{subfigure}
\usepackage{booktabs} % for professional tables

\usepackage{hyperref}
\usepackage{xpatch}
\usepackage[accepted]{icml2021}

\usepackage{kotex}
\usepackage{multirow}
\usepackage{comment}
\usepackage{amsmath}
\usepackage{mathtools}
\usepackage{adjustbox}

\DeclareMathOperator*{\argmax}{arg\,max}

\newcommand{\algofn}{\texttt}

\icmltitlerunning{Message Passing Adaptive Resonance Theory}

\begin{document}

\twocolumn[
\icmltitle{Message Passing Adaptive Resonance Theory \\ for Online Active Semi-supervised Learning}



\icmlsetsymbol{equal}{*}

\begin{icmlauthorlist}
\icmlauthor{Taehyeong Kim}{lge,snu}
\icmlauthor{Injune Hwang}{lge}
\icmlauthor{Hyundo Lee}{snu}
\icmlauthor{Hyunseo Kim}{snu} \\
\icmlauthor{Won-Seok Choi}{snu}
\icmlauthor{Joseph J. Lim}{lge,usc}
\icmlauthor{Byoung-Tak Zhang}{snu}
\end{icmlauthorlist}

\icmlaffiliation{lge}{AI Lab, CTO Division, LG Electronics, Seoul, Republic of Korea}
\icmlaffiliation{snu}{Seoul National University, Seoul, Republic of Korea}
\icmlaffiliation{usc}{University of Southern California, California, USA}

\icmlcorrespondingauthor{Byoung-Tak Zhang}{btzhang@bi.snu.ac.kr}

\icmlkeywords{Message Passing, Adaptive Resonance Theory, Online Learning, Active Learning, Machine Learning, ICML}

\vskip 0.3in
]



\printAffiliationsAndNotice{}  % leave blank if no need to mention equal contribution

%
%In this study, we propose \textit{Message Passing Adaptive Resonance Theory} (MPART) for online active semi-supervised learning.
%The proposed model learns the distribution and topology of the input data online.
%It then infers the class of unlabeled data and selects informative and representative samples through message passing between nodes on the topological graph.
%MPART queries the beneficial samples on-the-fly in stream-based selective sampling scenarios, and continuously improve the classification model using both labeled and unlabeled data.
%We evaluate our model with comparable query selection strategies and frequencies, showing that MPART significantly outperforms the competitive models in online active learning environments.
%Through message passing on the topological graph, the model actively selects informative and representative queries and infers the class of unlabeled data.

\begin{abstract}
Active learning is widely used to reduce labeling effort and training time by repeatedly querying only the most beneficial samples from unlabeled data.
In real-world problems where data cannot be stored indefinitely due to limited storage or privacy issues, the query selection and the model update should be performed as soon as a new data sample is observed.
Various online active learning methods have been studied to deal with these challenges;
however, there are difficulties in selecting representative query samples and updating the model efficiently without forgetting.
In this study, we propose \textit{Message Passing Adaptive Resonance Theory} (MPART) that learns the distribution and topology of input data online.
Through message passing on the topological graph, MPART actively queries informative and representative samples, and continuously improves the classification performance using both labeled and unlabeled data.
We evaluate our model in stream-based selective sampling scenarios with comparable query selection strategies, showing that MPART significantly outperforms competitive models.
\end{abstract}

\section{Introduction}
\label{introduction}

\begin{figure*}[ht]
\vskip 0.1in
\begin{center}
\centerline{\includegraphics[width=0.86\textwidth]{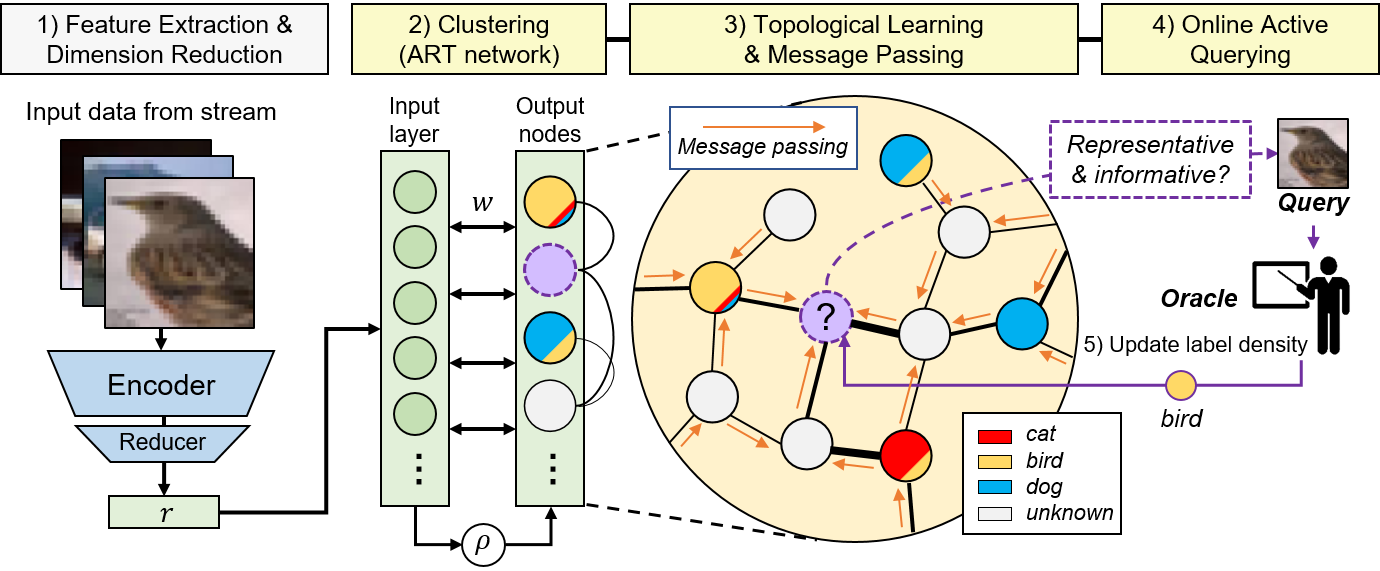}}
%\vskip -0.05in
\caption{Overview of Message Passing Adaptive Resonance Theory (MPART). The feature is extracted from the input sample, and then MPART continuously constructs a weighted graph based on ART network by learning distributions and topology of the input data. It uses a message passing method to infer the class label and estimate the uncertainty of the input sample for querying. A sample is queried according to the query selection strategy, and the label density of the topological graph is updated using the collected labels.}
%Note that all the processes are done online.}
\label{fig:mpart}
\end{center}
\vskip -0.2in
\end{figure*}

The recent success of deep learning in the field of visual object recognition and speech recognition is largely attributed to the massive amount of labeled data.
However, most of the data samples in the real-world are not labeled, so it takes a lot of time and effort to label and use them for deep learning.
In addition, in many countries, the collection and storage of medical data or data from personal robots are prohibited due to privacy concerns.
This burden impedes the widespread use of deep learning in real-world applications such as medical AI, home appliances and robotics, and separates the best solutions from becoming the best real-world solutions.

From this point of view, active learning is a promising field of machine learning.
%It reduces the labeling cost and effort by repeatedly selecting the most beneficial data among unlabeled set and requesting the oracle to label it.
The goal of active learning is to reduce the labeling cost by querying labels for only a small portion of the entire unlabeled dataset.
The query selection is made according to which samples are most beneficial, i.e. both informative and representative \cite{settles2009active}.
In most active learning algorithms, samples are selected from a large pool of data for querying \cite{settles2011theories, gal2017deep, sener2017active, beluch2018power}.
Once the queried data is labeled, it is accumulated in the training dataset and the entire model is trained again.
%This process is repeated, typically assuming that all the labeled and unlabeled data can be stored and the learner can access the data at any time again.
This process is repeated, typically assuming that all the data can be stored and the model can access the data again at any time.
In the aforementioned real-world problems, however, it is impossible to store a massive amount of data due to limited storage or privacy issues.
In addition, it is highly inefficient to repeatedly train the entire model with large data and review all the unlabeled data for querying whenever a new label is acquired.

In contrast, the active learning paradigm in an online manner does not assume that data samples can be accessed repeatedly but instead, the input data is given as a continuous stream \cite{lughofer2017line}.
This entails that uncertainty estimation of input samples and the decision of whether to query or not should be made online.
%The decision for querying and the training result with the sample affect the uncertainty of subsequent input samples.
The result of querying and training with one sample affects the uncertainty of subsequent input samples.
Therefore, the model update should be done on-the-fly so that the new uncertainty distribution is estimated for the next query.
In such scenarios, it is difficult to select a query sample that is representative as well as informative because all data samples cannot be reviewed at the same time.
%Moreover, because the labeling cost can be expensive and the oracle is not always available, the number of query may be limited \cite{hao2017second, zhang2018online, serrao2018active}.
Moreover, since the labeling cost is expensive and the oracle is often unavailable, the number of queries may be limited \cite{hao2017second, zhang2018online, serrao2018active}.
%Also, updating the model in such online learning can cause forgetting.
Catastrophic forgetting is another issue to overcome when learning data online \cite{lee2017overcoming}.
This online active learning process, while more suitable for real-world problems, is therefore more challenging than offline active learning.

%In this regard, we propose Message Passing Adaptive Resonance Theory (MPART) for online active semi-supervised learning.
%MPART learns the distribution of the input data and forms a topological graph online based on the ART network \cite{grossberg1987competitive} which keeps the existing knowledge stable when learning new data.
%Then it utilizes a message passing method on the graph to estimate the class label and uncertainty of the input sample in a semi-supervised manner.
%MPART gradually improves its performance as the data accumulates by online active learning.
%The proposed model can efficiently learn new data from continuous stream by updating only a part of the model without reviewing the data already learned.
%Figure \ref{fig:mpart} shows the overview of the proposed model.
In this regard, we propose Message Passing Adaptive Resonance Theory (MPART) that addresses these problems in online active learning.
MPART learns the distribution of the input data without forgetting based on ART \cite{grossberg1987competitive} which keeps the existing knowledge stable when learning new data.
%MPART's backbone network is based on ART \cite{grossberg1987competitive}, which allows continual learning of the input distribution without forgetting.
By utilizing a novel topology learning method on the learned distribution, MPART forms a topological graph using both labeled and unlabeled data.
%ART 기반의 topo learning 방법 제안 --> 포겟팅 없이 분포 학습
%MPART learns the topological structure from the input distribution and therefore is capable of selecting representative samples to query.
%그래서 대표적 샘플 쿼리가 가능하다.
%The underlying network based on ART \cite{grossberg1987competitive} continuously forms new concepts while keeping the existing ones.
%In order to exploit the limited amount of labels acquired, our model uses a message passing method to exchange class information within the topological graph.
%In order to exploit the limited amount of labels acquired, our model uses a message passing method on the topological graph, which compensates for the lack of class information by aggregating the information from a broader part of the graph.
In order to exploit the limited amount of labels acquired, our model uses a message passing method which compensates for the lack of class information by propagating the information within the topological graph.
Based on the learned topology, MPART can also select beneficial samples to query.
The entire learning process, depicted in Figure \ref{fig:mpart}, is performed in an online manner.
%The experimental results show that MPART can efficiently learn the data distribution from a continuous stream.
We evaluate our model in the online active learning environments, showing that MPART significantly outperforms competitive models.

The main contributions can be summarized as follows.
\begin{itemize}
\setlength\itemsep{0em}
\item We propose a novel method that learns the distribution of input data by continuously constructing a weighted topological graph based on the ART network.
\item We propose a message passing method for the graph to estimate the class labels in a semi-supervised manner and to select beneficial samples for querying.
\item We design an online active learning task where the frequency of query is limited and the total number of classes is unknown.
We validate the proposed model with various query selection strategies and datasets.
The results of the experiment show that our model significantly outperforms the competitive models in online active learning environments.
\end{itemize}

\section{Related Work}
\label{sec:related}
\textbf{Topology Learning. }
Topology learning aims to learn the structural properties and topological relations underlying the input data.
Additional knowledge such as class label or uncertainty can be obtained from the structural relationship using the topological information.
SOM \cite{kohonen1990self}, a type of artificial neural network, learns the topological properties of the input data using neighborhood function and competitive Hebbian learning (CHL).
This method is useful for dimension reduction of high-dimensional input data, but has a disadvantage in that it requires \textit{a priori} selection of a subspace.
GNG \cite{fritzke1995growing} and its derived models such as SOINN \cite{shen2006incremental} and E-SOINN \cite{shen2007enhanced} are incremental network models which can learn the topological relations of input data using the CHL.
There are also variants of ART networks (see Section \ref{sec:art} for details) which integrate new knowledge into its existing knowledge so that what has already been learned is not forgotten by the new learning.
Fuzzy ART-GL \cite{isawa2007fuzzy} uses Group Learning that creates connections between similar categories. 
TopoART \cite{tscherepanow2010topoart} combines incremental clustering with topology learning, which enables stable online clustering of non-stationary input data.
%ART will be introduced briefly in Section \ref{sec:art}.

\textbf{Semi-supervised Learning. }
Semi-supervised learning uses a small amount of labeled data with a large amount of unlabeled data to improve the performance of the model.
\cite{iscen2019label} and \cite{douze2018low} use the label propagation method to assign labels for unlabeled data using the nearby labeled data.
GAM \cite{stretcu2019graph} uses an agreement model that calculates the probability of two nodes sharing the same label on the graph.
%EGNN \cite{kim2019edge} predicts the edge-labels of a graph rather than the node-labels in order to estimate the similarity of node-labels.
EGNN \cite{kim2019edge} predicts the edge-labels of a graph to estimate the node-labels which are hard to be inferred directly in few-shot classification tasks.
%Batch SS-SOM \cite{braga2020deep} utilize SOM .
%However, most of these methods are not suitable for online learning, because they need predefined topological information or whole training data repeatedly.
However, most of these methods are not suitable for online learning, because they need predefined topological information or repeated usage of whole training data.
LPART \cite{kim2020label} uses online label propagation on the ART network trained in a semi-supervised manner to overcome this issue, but the information conveyed between the nodes is limited due to insufficient topological information.
MPART, on the other hand, addresses these problems by explicitly learning the topology of the input data.
%because it can't learn the topology of input data explicitly.

%Whereas MPART can effectively exchange meaningful info utilizing explicit topology. learn the data distribution from a continuous stream.
%not sufficiently learn and use topology information of input data

\textbf{Online Active Learning. }
The goal of online active learning is not only to reduce annotation costs, but also to continuously expand existing knowledge by exploring new information.
%OASIS \cite{goldberg2011oasis} is a Bayesian model using particle filtering to estimate the posterior.
OASIS \cite{goldberg2011oasis} is a Bayesian model that combines online active learning and semi-supervised learning methods.
SOAL \cite{hao2017second} and OA3 \cite{zhang2018online} utilize second-order information for online binary classification under limited query budgets.
%These are successful in binary classification, but the fixed number of classes makes it hard to adapt to increasing complexity of data.
While these methods are successful in binary classification, they cannot handle the increasing diversity of classes due to the assumption of a fixed number of classes.
%These methods are successful in binary classification, but the fixed number of classes makes it hard to adapt to various number of classes.
%Other methods \cite{loy2012stream, weigl2016improving} use ensemble models to compute the uncertainty of inputs and sample the model parameters to update the classifier.
%But the sampling in a complex task is difficult and inefficient, so these algorithms require the underlying model to be simple.
Other methods \cite{loy2012stream, weigl2016improving} employ consensus of models to better estimate uncertainty.
However, using multiple models incurs large computational costs, making them less scalable for complex tasks.
%To avoid this, we use a deterministic approach to calculate the uncertainty of data for active learning.
%The most closely related work to ours is \cite{shen2011incremental}, which addresses a very similar problem using Active learning SOINN (A-SOINN).
%It shows the capability of online active semi-supervised learning, however, the generated query might be ambiguous since it is not the input itself but the prototype of inputs (i.e. the representative value of the cluster).
%Moreover, this model stores only a single label in a cluster of inputs, so it cannot handle the possible mixture of classes within the cluster and is susceptible to noise.
The most closely related work to ours is \cite{shen2011incremental}, enhancing the SOINN to enable online active semi-supervised learning.
The proposed A-SOINN, however, selects the queries among representative values of clusters, which might be difficult for the oracle to recognize.
Moreover, since this model stores only a single label in a cluster, it cannot handle the mixture of classes within the cluster and is susceptible to noise.
%Our model solves these problems by querying the input instantly and using the label density of each node which is estimated via a message passing method.
Our model solves these problems by querying the input itself and using a distribution for labels in each node.

%Each related work shows a high potential in a variety of research fields to solve real-world problems, and motivates our idea of online active semi-supervised learning.
%In the following sections, we briefly introduce the ART network and our motivations, then a novel topology learning method based on FuzzyART \cite{carpenter1991fuzzy} is presented, followed by the class inference method using message passing, and the active querying process.

\section{Background}
\label{sec:art}
Each related work shows a high potential to solve real-world problems in various domains, and motivates our idea of online active semi-supervised learning.
In this section, we briefly introduce the ART networks and our motivations.

\subsection{ART Networks}

Adaptive Resonance Theory (ART), inspired by brain information processing mechanisms, is an unsupervised learning method for pattern recognition \cite{grossberg1987competitive}.
The basic principle of ART is that object recognition is achieved by the interaction of `bottom-up' sensory information and `top-down' expectations.
We refer to any neural network model based on ART as \textit{ART network}.
ART networks integrate new knowledge into the entire knowledge so that what has already been learned is not forgotten by new learning.
%A basic ART network consists of two fully connected layers responsible for inputs and outputs.

A basic ART network consists of two layers fully-connected to each other: an input layer and an output layer.
%The input vector is stored in the form of a prototype in a node of the output layer.
%When a one-dimensional feature vector is input to the ART network, the closest matching node in the output layer is selected as a winner node using the choice function.
As an input vector enters the input layer, it activates each node in the output layer according to the connection weight.
Using the choice function with the activation values, the closest matching node is selected as a winner.
%Then, the fitness of the input vector for the winner node is calculated using the match function, and if the value is greater than a vigilance parameter $\rho$, the input vector is integrated into the winner node.
Then, the fitness of the input to the winner node is calculated using the match function.
If the fitness is greater than a vigilance parameter $\rho$, the input vector is integrated into the winner, updating the associated weight.
%Here, if the value of the match function is less than $\rho$, a new node with the input vector as its prototype is created.
Otherwise, a new node is created with a weight set equal to the input vector.
With these mechanisms, ART networks can categorize incoming data online without forgetting.
There are various ART networks such as Fuzzy ART \cite{carpenter1991fuzzy}, Gaussian ART \cite{williamson1996gaussian} and Hypersphere ART \cite{anagnostopoulos2000hypersphere}.
We use Fuzzy ART as our backbone model, which incorporates fuzzy logic to enhance generalizability.
Fuzzy ART will be described in Section \ref{sec:topology} along with MPART's topology learning method.

%It does not require already learned data for retraining the model, and it can efficiently learn new data by updating only a part of the model.
%ART, a type of neural network, is an unsupervised learning method inspired by brain information processing mechanisms.
%When a sample is inputted into the network, ART checks whether it fits into one of the already stored clusters.
%If it fits then the sample is trained on the best matching cluster, otherwise a new cluster is created.
%This method provides a great advantage for online active learning because it does not require already learned data for retraining the model, and it can efficiently learn new data by updating only a part of the model.
% 김현서
%ART는 unsupervised 하게 feature 본연의 특징으로만 categorization, clustering을 진행하는 network다. 
%두 개의 layer(module? field? network?)가 상호작용하며 feature cluster 또는 node의 boundary? 크기?를 결정짓는다. 
%input이 들어오면 bottom layer에서 이 input이 기존 노드 중 어느 것과 가장 유사한지 결정하는 choice function을 계산한다. 
%그러면 Top down layer는 현재 input에서 Best match 노드가 “pay attention”/ expect할 수 있는 정도가 input의 어느정도 proportion인지 match function으로 계산하여서 그 정도가 vigilance parameter보다 작으면, 즉 input이 너무 novel하면  search를 다시 시작하거나 새로운 category를 만든다.
%vigilance parameter보다 클 경우 input을 best match node에 통합한다.\ref{sec:topology}.

\subsection{Motivations}
Due to the nature of ART, the ART networks can efficiently learn new data by updating only a part of the model.
%In addition, it does not need to predefine the structure and size of the model, which can self-organize irregular and complex data distributions.
In addition, it does not need to predefine the structure and size of the model, which allows flexible self-organization of irregular and complex data distributions.
%These are important properties in online learning, so we chose ART as a backbone model of MPART.
These properties play an important role in online learning, so we chose an ART network as a backbone model of MPART.

Moreover, in online active learning scenarios where data cannot be stored, it is difficult to select a representative sample for querying because all data samples cannot be reviewed simultaneously.
We address this problem by learning the topology of input data.
There is also an ART network called TopoART that can learn the topology.
However, TopoART, in which only the winner and the runner-up node are connected, cannot effectively represent the relationships between nodes.
In TopoART, the strength of the edge continuously increases as the data is learned, so a proper normalization method is required to use the edge information for message passing on the topological graph.
Therefore, we propose a novel method to learn the topology and utilize it for message passing to select queries and predict labels.

\section{Methods}
MPART learns the distribution and topology of input data to construct a topological graph online.
Then, class prediction and uncertainty estimation of the input data are performed using a message passing method on the graph.
The estimated uncertainty is used to select and query useful samples.
The entire process of MPART is described in Algorithm \ref{alg:mpart}.
In the following sections, we describe the topology learning, message passing, and active querying methods.

\subsection{Topology Learning}
\label{sec:topology}
\textbf{Node Formation. }
%We first extract $h$-dimensional representation vector $r_t \in [0,1]^{h}$ from the input data $x_t$ using the pre-trained BYOL \cite{grill2020bootstrap} and Parametric UMAP \cite{sainburg2020parametric} (see Section \ref{sec:featext} for details).
We first extract representation vector $r_t \in [0,1]^{n_r}$ from the input data $x_t$ using the pre-trained BYOL \cite{grill2020bootstrap} and Parametric UMAP \cite{sainburg2020parametric} (see Section \ref{sec:featext} for details).
As in Fuzzy ART, the representation vector $r_t$ is complement coded to $I_t = [r_t, \vec{1} - r_t]$ in order to avoid proliferation of prototypes \cite{carpenter1991fuzzy}.
With this, we can measure the similarity between the input and the category node $j$ using the match function $M_j$ and the choice function $T_j$ as follows.
\begin{equation} \label{eq:activation}
M_j(I_t) = \frac{\left\| I_t \wedge w_j \right\|_1}{\left\| I_t \right\|_1}, \qquad
T_j(I_t) = \frac{\left\| I_t \wedge w_j \right\|_1}{\alpha + \left\| w_j \right\|_1}
\end{equation}
Here, $\wedge$ is the element-wise minimum operator, $\| \cdot \|_1$ is the L1 norm, and $\alpha > 0$ is a choice parameter.
A node $j$ becomes \textit{activated} by $I_t$ if $M_j(I_t)$ is greater than or equal to a vigilance parameter $\rho \in (0, 1)$.
%If none is activated, a new node is created with $I_t$ as an initial weight $w$ and a winning count $d$ of 1.
When there are multiple activated nodes, or \textit{co-activated nodes}, one with the highest $T_j(I_t)$ is chosen as a \textit{winner} denoted by $J_t$.
In case only one node is activated, we also refer to that node as the winner $J_t$.
We update $w_{J_t}$ with a learning rate $\beta \in (0, 1]$, and increase $d_{J_t}$ by 1 as follows.
\begin{equation} \label{eq:update}
\begin{gathered}
w_{J_t}^{new} = \beta (I_t \wedge w_{J_t}^{old}) + (1 - \beta) w_{J_t}^{old} \\
d_{J_t}^{new} = d_{J_t}^{old} + 1
\end{gathered}
\end{equation}
If none is activated, a new node $J_t$ is created with an initial weight $w_{J_t}=I_t$ and a winning count $d_{J_t}=1$.

% In addition to the node formation based on basic ART, edges between the co-activated nodes are developed which grants topological properties without disrupting ART characteristics.
\textbf{Edge Formation. }
In addition to the node formation, edges between the co-activated nodes are developed.
Unlike SOINN or TopoART in which only the winner and the runner-up node are connected, we connect the winner to every co-activated node to better represent more complex topology of multiple nodes.
For an edge connecting nodes $i$ and $j$, its count $c_{ij}$ is set as the number of times $i$ and $j$ have been co-activated.
The edge weight $e_{ij}$ is defined as a ratio of $c_{ij}$ to the sum of winning counts of incident nodes $i$ and $j$ as shown in Equation \ref{eq:edge_weight}.
\begin{equation} \label{eq:edge_weight}
e_{ij} = \frac{c_{ij}}{d_i + d_j}
\end{equation}
%
%The value of $e_{ij}$ approaches to 1, which is the upper bound, when the two nodes have always been co-activated whenever either one of them was the winner.
%It allows the amount of information transmitted between two nodes to be appropriately adjusted according to the edge weight.
%Therefore, the edge weight $e_{ij}$ is a good indication of the relationship between the two nodes, regardless of the edge weight with the other nodes.
The weight $e_{ij}$ well describes the correlation between $i$ and $j$.
If the two nodes have never been co-activated, $e_{ij}$ is 0.
As they get co-activated more frequently, $e_{ij}$ grows larger; it reaches the maximum value of 1 if $i$ and $j$ have been co-activated whenever one of them was the winner.
Since $e_{ij}$ is bounded to $[0, 1]$, it can be used to adjust the degree of message passing without normalization.
%It allows the amount of information transmitted between two nodes to be appropriately adjusted according to the edge weight.
The edge formation does not interrupt the node formation, so the properties of the underlying ART network are also preserved.
%Introducing the development of edges does not impede original ART properties.
%As a result, this process does not need to pre-define the structure and size of the network, and exploits the self-organizing property to learn the topology of data with irregular and complex distributions.
%It can also be operated online without catastrophic forgetting and has low computational costs.
%The example results of the topology learning are shown in Figure \ref{fig:topology_visualization}.

\begin{algorithm}[t]
\caption{The MPART algorithm}
\label{alg:mpart}
\begin{algorithmic}
    %\COMMENT{We used a complete graph for simplicity of notations.}
    \STATE $V \leftarrow \{\}, C \leftarrow \{\}$ %\COMMENT{initialize complete graph $G(V)$ and known class set $C$}
    \FOR{$x_t$ {\bfseries in} input data stream}
        \STATE $r_t \leftarrow \algofn{DimensionReduction}(x_t)$ %\COMMENT{see Section \ref{sec:featext} for details}
        \STATE $I_t \leftarrow [r_t, \overrightarrow{1} - r_t]$ %\COMMENT{complement coding}
        \STATE $A \leftarrow \{\}$
        \FOR{$j$ {\bfseries in} $1, \ldots, |V|$}
            \STATE $M_j \leftarrow \|I_t \wedge w_j\|_1 \; / \; \|I_t\|_1$ %\COMMENT{compute match function}
            \STATE $T_j \leftarrow \|I_t \wedge w_j\|_1 \; / \; (\alpha + \|w_j\|_1)$ %\COMMENT{compute choice function}
            %\IF[find co-activated nodes]{$M_j \geq \rho$}
            \IF{$M_j \geq \rho$}
                \STATE $A \leftarrow A \cup \{j\}$
            \ENDIF
        \ENDFOR

        \IF{$A$ is empty}
            \STATE $J_t \leftarrow |V| + 1, \quad V \leftarrow V \cup \{J_t\}$ %\COMMENT{create new node}
            \STATE $c_{J_{t}v} \leftarrow 0,\; c_{vJ_{t}} \leftarrow 0 \quad \forall v \in V - \{J_t\}$ %\COMMENT{initialize edge counts}
            \STATE $q_{J_t}(y) \leftarrow 0 \quad \forall y \in C$ %\COMMENT{initialize label density function}
            \STATE $w_{J_t} \leftarrow I_t, \quad d_J \leftarrow 1$ %\COMMENT{initialize weight and winning count}
        \ELSE
            \STATE $J_t \leftarrow \argmax_{j \in A}(T_j)$ %\COMMENT{find winner}
            \STATE $c_{J_{t}v} \leftarrow c_{J_{t}v} + 1,\; c_{vJ_{t}} \leftarrow c_{vJ_{t}} + 1 \quad \forall v \in A - \left\{J_t\right\}$ \\
            \STATE $w_{J_t} \leftarrow \beta (I_t \wedge w_{J_t}) + (1 - \beta) w_{J_t}$ %\COMMENT{update weight}
            \STATE $d_{J_t} \leftarrow d_{J_t} + 1$ %\COMMENT{update winning count}
        \ENDIF

        \STATE $q_{J_t}^{(L)}, d_{J_t}^{(L)} \leftarrow \algofn{MessagePassing}(J_t, c, d, q)$ %\COMMENT{see Section \ref{sec:message_passing} for details}
        \STATE $p_{t}, \hat{y} \leftarrow \algofn{NodeClassification}(q_{J_t}^{(L)})$ %\COMMENT{see Section \ref{sec:message_passing} for details}
        \STATE $s_{t} \leftarrow \algofn{UncertaintyEstimation}(p_{t}, q_{J_t}^{(L)}, d_{J_t}^{(L)})$ %\COMMENT{see Section \ref{sec:query} for details}
        %\IF[see Section \ref{sec:strategy} for details]{$u_{J}$ satisfies query condition}
        \IF{$s_{t}$ satisfies query condition}
            \STATE $y_t \leftarrow \algofn{QueryLabel}(x_t)$
            \STATE $q_{J_t}(y_t) \leftarrow q_{J_t}(y_t) + 1$ %\COMMENT{initial $q_J^{(0)}(y_i)$ value is 0}
            \STATE $C \leftarrow C \cup \{ y_t \}$ %\COMMENT{add new label to known class set $C$}
        \ENDIF
    \ENDFOR
\end{algorithmic}
\end{algorithm}

\subsection{Message Passing}
\label{sec:message_passing}
%MPART performs message passing between nodes through a gradually formed topological graph.
%In a graph network, message passing is a method that can predict the state of target nodes based on their neighboring nodes.
%%By using this method, Graph Neural Networks (GNN) could update the representation of the target node according to the neighboring nodes \cite{hamilton2017representation}.
%%This makes it possible to predict each node's state more accurately based on the overall graph.
%In this study, message passing for MPART is defined as Equation \ref{eq:aggregation} for node classification and uncertainty estimation.
Nodes formed in MPART possess additional information vectors such as class probabilities or uncertainty measures for better inference and querying.
%To enhance the quality of information, we use message passing to aggregate the global information from neighboring nodes.
To compensate for the lack of information, we use message passing to aggregate the information within the topological graph.
Before describing the procedure, we first define a notion of \textit{neighbors}.
Given nodes $i$ and $j$, if the shortest path from $i$ to $j$ exists and its length is $l$, then we say $j$ is an $l$-hop neighbor of $i$.
Any node is a 0-hop neighbor to itself.
A set of all $l$-hop neighbors of $i$ is denoted as $\mathcal{N}_i^{(l)}$; a set of direct neighbors $\mathcal{N}_i^{(1)}$ will also be written as $\mathcal{N}_i$.
We further denote an $l$-hop neighborhood $\mathcal{N}_i^{(0:l)}$ as a union of $\mathcal{N}_i^{(0)}, \ldots, \mathcal{N}_i^{(l)}$.

To illustrate message passing, let $X$ denote an information vector (or scalar) of interest such as winning count $d$ or label density $q$ (will be described later).
%, i.e. $X$ can be substituted with $d$, $q$, or any other quantity.
A single-layer message passing on the target node $i$ updates $X_i$ by adding $X_j$ of every neighboring node $j \in \mathcal{N}_i$, weighted by $e_{ij}$ and discounted by a propagation rate $\delta \in [0, 1]$.
%We can achieve 2-layer message passing on a target node $i$ by first updating the $X$'s in $\mathcal{N}_i^{(0:1)}$ then carrying out another update on carrying out another update on $i$.
%For a target node $i$, we can achieve 2-layer message passing by first updating the $X$'s in $\mathcal{N}_i^{(0:1)}$ then carrying out another update on $i$.
Multi-layer message passing is achieved by performing the single-layer message passing recursively, aggregating the information from broader neighboring area.
%$l$ consecutive updates will lead to an $l$-layer message passing, aggregating information from broader neighboring area.
The recursive procedure for $L$-layer message passing targeted at the winner $J_t$ can be formalized as Equation \ref{eq:aggregation}, where $X_{i}^{(l)}$ denotes the $l$-layer aggregated information and $X_{i}^{(0)}$ is set as $X_{i}$.
\begin{equation} \label{eq:aggregation}
\begin{gathered}
X_{i}^{(l)} = X_{i}^{(l-1)} + \delta \sum_{j \in \mathcal{N}_i} e_{ij} X_{j}^{(l-1)}, \ \forall i\in \mathcal{N}_{J_t}^{(0:L-l)}
\end{gathered}
\end{equation}
We finally obtain the $L$-layer aggregated information $X_{J_t}^{(L)}$ by applying this update for $l$ from 1 to $L$.
Note that we only have to consider nodes in $\mathcal{N}_{J_t}^{(0:L-l)}$ when performing the $l$-th layer update since the information outside this set will not be propagated to $J_t$ within the remaining $(L-l)$ updates.
%Note that we only have to consider nodes in $\mathcal{N}_{J_t}^{(0:L-l)}$ when performing the $l$-th layer update since the information on nodes excluded from this set will not be propagated to $J_t$ within the remaining $(L-l)$ updates.
This reduces the computational cost compared to when reviewing all the nodes in the graph.
%Here, $X_{i}$ and $X_{j}$ are information vectors of interest, such as class probability, on the target and neighboring nodes, and $\mathcal{N}_i$ is a set of all neighbors of node $i$.
%$J_t$ is a winning node with given input $x_t$ and $\mathcal{N}_{J_t}(L-l)$ is $(L-l)$-hop neighborhood of node $J_t$.
%$\delta$ is a constant between 0 and 1 to determine the propagation rate.
%By performing this process on all node information $X_{i}^{(l)}$ of the graph, the updated node information $X_{i}^{(l+1)}$ for the next layer $(l+1)$ is obtained.
%The base value $X_{i}^{(0)}$ for layer $0$ is the node's own value $X_{i}$.
%The more repeatedly this method is performed on multiple layers, the broader aggregation is possible for information on the nodes that are further away from the reference node.
%Finally, we can use the node information $X_{i}^{(L)}$ of the final layer $L$ to perform the task we want.

%%This process can be performed on all nodes in parallel, because the node information of layer $(l+1)$ depends only on the node information of the previous layer $l$.
%The calculation of all the information vectors $X^{(l+1)}$ at layer $(l+1)$ referring to layer $l$ can be performed in parallel.
%In addition, it allows efficient computation as it only updates a partial area that needs to be newly calculated according to the number of layers and input sample.

\textbf{Node Classification. }
MPART infers the class of input $x_t$ by aggregating the class information near the winner $J_t$.
The class information $q_i$ of a node $i$, called the label density, is a distribution over a set of known class labels $C$ and indicates how probable the node belongs to each class.
We use distributions rather than single values to cope with situations where one node represents samples of different classes.

Before receiving any label, each $q$ is set to an empty distribution, i.e. an empty vector.
As a new label is received and added to the label set $C$, every $q$ is expanded by one dimension (zero-valued) which will be responsible for that label.
For new nodes, label densities are set to zero vectors of dimension $|C|$.
%The value of $q$ is updated by direct labeling or message passing.
When a label $y_t$ is provided with an input $x_t$, the corresponding density value in the winner ${J_t}$, i.e. $q_{J_t}(y_t)$, increases by 1.
This increment also applies if the label was obtained via active querying.
Since labels are provided rarely, we perform $L$-layer message passing for $q$ to alleviate the deficiency.
The aggregated label density $q_{J_t}^{(L)}$ contains class information of neighboring nodes, so it is representative and robust to noise.
We obtain the class probability distribution $p_t$ by normalizing it as Equation \ref{eq:classification}.
%If $q_{J_t}^{(L)}$ is a zero vector then we assume that $p_t$ is equally distributed.
%
\begin{equation} \label{eq:classification}
p_{t}(y) = \dfrac{q_{J_t}^{(L)}(y)}{\sum_{y' \in C}q_{J_t}^{(L)}(y')}
\end{equation}
If $q_{J_t}^{(L)}$ is a zero vector, we set $p_t$ to a uniform distribution.
The label of the input is inferred as one with the highest probability.
This distribution is also used to estimate the label uncertainty of a node, as described in the next section.

\subsection{Active Querying}
\label{sec:query}
 
\begin{figure}[t]
\vskip 0.05in
\begin{center}
%\centering
    \centerline{\includegraphics[width=0.95\columnwidth]{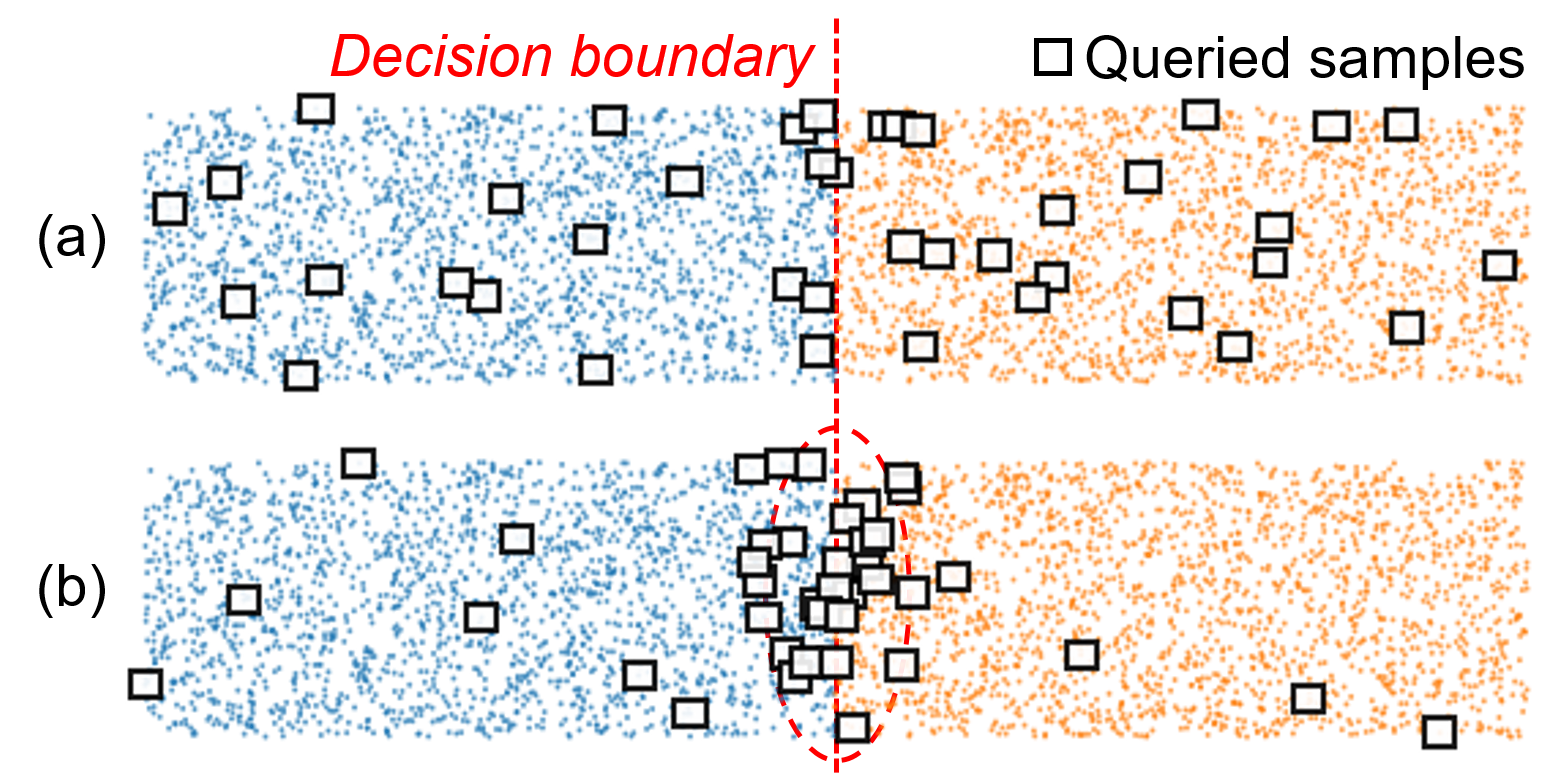}}
    %\vskip -0.1in
    \caption{Example queries (black boxes) selected online according to the query selection score, for a simple binary classification of 2,000 samples each (blue and orange). (a) The first 40 queries are distributed relatively evenly, while (b) the latter 40 queries are concentrated around the decision boundary.}
    %\caption{We performed a simple binary classification task to investigate the distribution of queried samples according to the learning progress. The colored dots visualize the input data and their labels, and black boxes indicate the queried samples. For a total of 4000 input data, 40 queries were selected online by MPART. (a) and (b) visualize selected queries of the first 20 and the latter 20, respectively. (a) Queries selected at the beginning of training are distributed relatively evenly, while (b) queries selected at the end of training are mostly distributed around the decision boundary according to the density-weighted query selection scores $s_t$.}
\label{fig:decision_boundary}
\end{center}
\vskip -0.2in
\end{figure}

%We use both uncertainty-based sampling and density-weighted method for query selection.
%Density-weighted methods are less prone to query outliers than uncertainty-based sampling \cite{settles2009active}.
%We explicitly estimate the distribution of the input and use it for query selection.
We combine the uncertainty-based sampling with density-weighted method for query selection.
The uncertainty is measured per node using the label density and weighted by distribution density of data to produce the selection score.
This score can be used with various selection strategies, including those suggested in Section \ref{sec:strategy}.

\textbf{Uncertainty Estimation. }
%We use two different metrics to measure the uncertainty of the nodes as in \cite{kim2020label}.
%The first uncertainty $u_{e}$ which can be seen as an epistemic uncertainty \cite{hullermeier2021aleatoric} is based on the quantitative information of label density $q_{J_t}^{(L)}$ of winning node as shown in Equation \ref{eq:epistemic}.
%Here, $k_e$ is a positive constant for sensitivity.
%This uncertainty has a high value in the region lacking labels in the input data distribution.
We use two kinds of uncertainty measures as in \cite{kim2020label}.
The first uncertainty $u_e$ indicates the insufficiency of class information accumulated in each node.
It is determined by the sum of label density values over all known classes.
As more labels are input or more times of message passing are performed, $u_e$ decreases as the sum of densities increases.
For an input sample $x_t$, $u_e$ is defined as that of the winner $J_t$ as in Equation \ref{eq:epistemic}, with a sensitivity constant $k_e > 0$ and $\tanh$ for clamping.
\begin{equation} \label{eq:epistemic}
u_{e} = 1 - \tanh\left(k_e \sum_{y \in C} q_{J_t}^{(L)}(y)\right)
\end{equation}
%
%The second uncertainty, $u_{a}$, can be seen as an aleatoric uncertainty \cite{hullermeier2021aleatoric} which is measured by the entropy of the classification probability as shown in Equation \ref{eq:aleatoric}.
%This uncertainty tends to have a high value near the decision boundary, which can help in selecting samples near the boundary.
The second uncertainty $u_{a}$ measures the class impurity, i.e. how many classes are mixed within a node.
We can use the normalized entropy of class probability, while setting $u_{a}$ to 0 when no more than one label is observed.
\begin{equation} \label{eq:aleatoric}
u_{a} =
\begin{cases}
\dfrac{-\sum_{y \in C} p_{t}(y)\log p_{t}(y)}{\log(|C|)} ,& \text{if } |C| > 1 \\
0,& \text{otherwise}
\end{cases}
\end{equation}

The two quantities, $u_e$ and $u_a$, can be viewed as epistemic and aleatoric uncertainties respectively, in that $u_e$ results from insufficient observation while $u_a$ arises due to ambiguity in the input data.
%The two uncertainties are complementary for query selection strategies.
These uncertainties are complementary when used for query selection, since $u_e$ plays an important role in the early stage of learning while $u_a$ is more crucial after enough number of labels have been acquired.
%Therefore, we put these two uncertainty measures together with a weight $\tau$ between 0 and 1 to define a query selection score $u_{t}$ as in Equation \ref{eq:u_sum}.
We combine them with a weight $\tau \in [0, 1]$ to get a query selection score $u_{t}$ as shown in Equation \ref{eq:u_sum}.
%If the label density of nodes is high enough, $u_{e}$ converges to 0, so $u_{t}$ changes according to $u_{a}$ and more data near the decision boundary is selected as a query target as shown in Figure \ref{fig:decision_boundary}.
Figure \ref{fig:decision_boundary} illustrates that the distribution of queries selected using $u_t$ tends to be sparse at first, but concentrated near the decision boundary as the learning progresses.
\begin{equation} \label{eq:u_sum}
u_{t} = \tau u_{e} + (1-\tau) u_{a}
\end{equation}

\begin{figure}[t]
%\vskip 0.2in
\centering
%\begin{center}
    \centerline{\includegraphics[width=0.87\columnwidth]{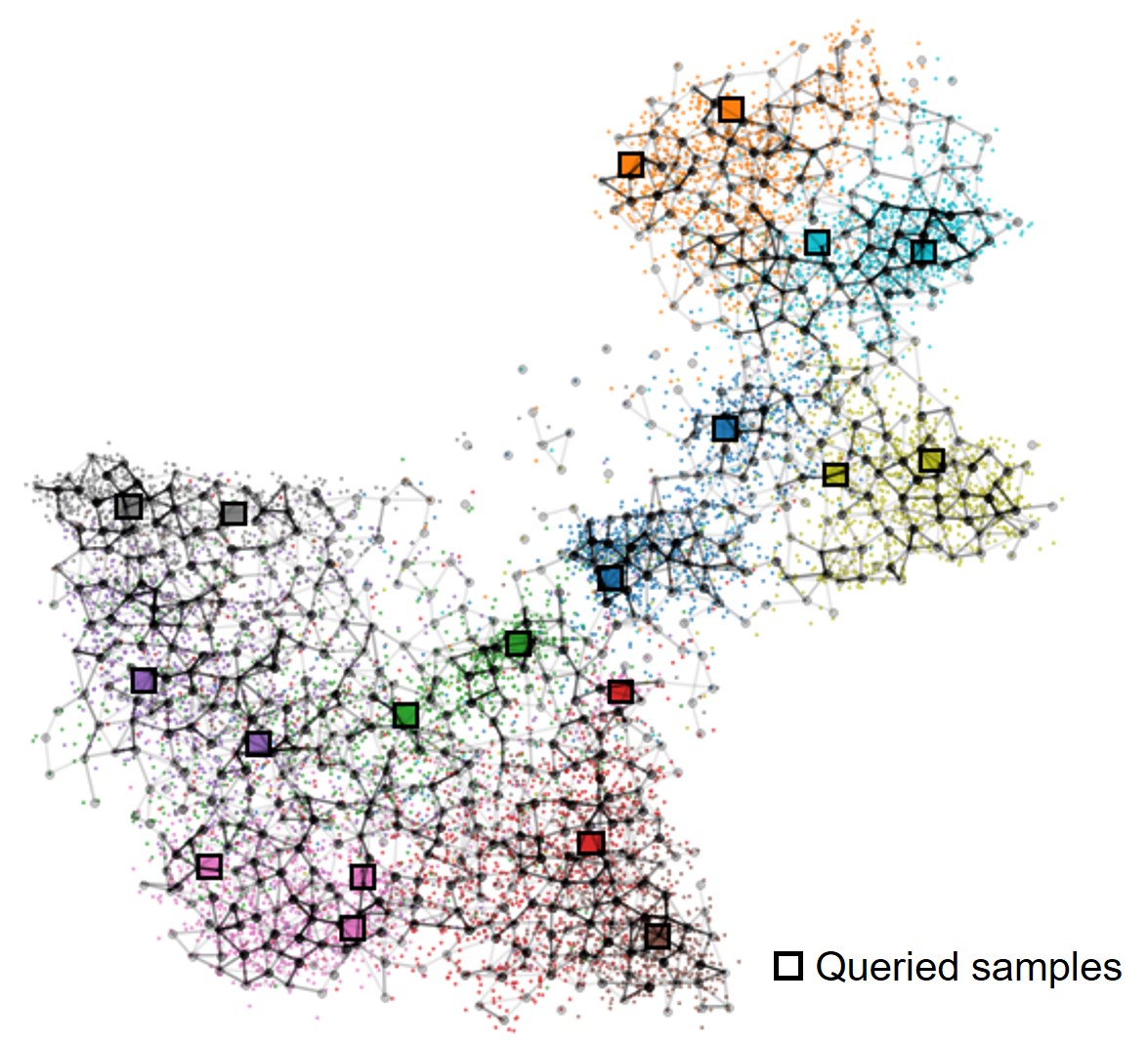}}
    %\vskip -0.1in
    \caption{The visualization of result on the CIFAR-10 dataset for 1/500 query frequency and $L=3$ using `Explorer' strategy. The color of the dots represent the label of input data, and the intensity of the topological graph represents the density of nodes and edges. The distribution of the queried samples is spread evenly.}
\label{fig:topology_visualization}
%\end{center}
\vskip -0.1in
\end{figure}

\textbf{Density-Weighted Method. }
Beneficial samples should be representative as well as informative \cite{huang2014active}.
Both uncertainties $u_{a}$ and $u_{e}$ are considered representative because they are derived from label density $q_{J_t}^{(L)}$ aggregated from surrounding nodes.
%This is a desired property of criteria for selecting beneficial queries \cite{huang2014active}.
Additionally, we aggregate the winning count $d_i$ to select representative samples, which represents the distribution density of the input data.
%but those two are complementary in uncertainty estimation.
%$\tau$ is a parameter between 0 and 1 for the weight of $u_{J,a}$ and $u_{J,e}$.
Finally, the density-weighted query selection score $s_{t}$ using data density $d_{J_t}^{(L)}$ is defined as in Equation \ref{eq:score}.
\begin{equation} \label{eq:score}
\begin{gathered}
d_{i}^{(l)} = d_{i}^{(l-1)} + \delta \sum_{j \in \mathcal{N}_i} e_{ij} d_{j}^{(l-1)}, \ \forall i\in \mathcal{N}_{J_t}^{(0:L-l)} \\
s_{t} = \tanh\left( k_d \cdot d_{J_t}^{(L)} \right) \cdot u_{t}
\end{gathered}
\end{equation}
Here, $k_d$ is a positive constant for sensitivity.
If the density-weighted query selection score $s_{t}$ of the input $x_t$ satisfies the condition according to the query selection strategy (see Section \ref{sec:strategy} for details), the model immediately queries the oracle to get a label $y_t$.
An example of training results using the CIFAR-10 dataset is visualized in Figure \ref{fig:topology_visualization}.

\section{Experiments}
\label{sec:experiment}

%We investigate the effectiveness of the proposed model in online active learning scenarios.
We investigated the effectiveness of MPART in online active learning scenarios.
%To do this, we designed a task described in Section \ref{sec:task} and evaluated the performance of the model in various combinations of task and model settings.
To do this, we designed a task described in Section \ref{sec:task} and evaluated the performance of the model with various settings.
%We also compared the performance of the proposed model to two other competitive models.
We also compared the performance to that of two competitive models: LPART \cite{kim2020label} and A-SOINN \cite{shen2011incremental}.
%We repeated each experiment 30 times to obtain statistical significance.
For statistical significance, we repeated each experiment 30 times.

%The parameter $\delta$ for message passing in Equations \ref{eq:classification} and \ref{eq:score} was set to 0.1, and the parameter $k_e$, $\tau$ and $k_d$ in Equations \ref{eq:epistemic}, \ref{eq:u_sum} and \ref{eq:score} were empirically set to 1.0, 0.7 and 0.01 respectively.
The propagation rate $\delta$ for message passing was set to 0.1, and the parameters $k_e$, $\tau$ and $k_d$ used for the score calculation were set to 1.0, 0.7 and 0.01, respectively.
%Please see the Appendix for all other model parameters.
For other parameter settings, please refer to the Appendix.

%%% TASK DESCRIPTION %%%%
\subsection{Task}
\label{sec:task}
%In many applications such as medical diagnosis and malicious URL detection \cite{bhattacharjee2017prioritized}, the annotations are often time-consuming and expensive.
%In these environments, only a few annotations may be possible during a specific time period.
%In addition, data storage may be restricted due to privacy concerns.
%To deal with these problems, an online active learning task is proposed, which limits the number of queries for a given time period of continuous data streaming \cite{attenberg2011online}, unlike typical active learning tasks with a limited total query budget.
We set up an online active learning task to imitate real-world scenarios where only a few number of annotations are possible and the data cannot be stored.
There is a previous study that has proposed this kind of task for stream-based selective sampling \cite{attenberg2011online}.
In this previous work, the query budget is set as the maximum number of queries allowed in a fixed input period instead of the total number of queries throughout the training.
Similarly, we constrained the model to query for certain number of times within each period, assuming that the oracle can annotate only a few samples in a period.

%Similarly, we designed an online active learning task with a fixed query budget per time period to evaluate the performance of the proposed model.
%The basic goal is to correctly classify the given data, but the number of possible classes is not known in advance.
%The whole training dataset is initially unlabeled and the model can inquire the oracle for ground truth labels.
%In this task, the query period and query budget per period is limited to imitate real-world scenarios where labeling is expensive.
%Specifically, we only allow $B$ queries for every $W$ consecutive samples, where $B$ is 1 or 2, and $W$ is one of the four fixed periods: 100, 500, 1,000 and 2,000.
%For simplicity, we denote the \textit{query frequency} as $B$/$W$.
%All input samples are provided online, i.e. one after another, and cannot be collected in storage for model retraining.
The basic goal of this task is to perform multi-class classification, where the number of classes is unknown in advance.
The input samples are provided sequentially as a stream, i.e. one after another, and cannot be stored or reused.
The whole training dataset is initially unlabeled and the model can inquire the oracle about the labels of a small number of inputs.
Specifically, the number of queries is limited to a fixed budget $B$ within a period of $W$ consecutive inputs.
We denote such constraint as a query frequency $B/W$.
The final classification performance is evaluated on the hold-out test dataset.

%For this task, we used four kinds of datasets with different distributions: Mouse retina transcriptomes \cite{macosko2015highly, polivcar2019opentsne}, Fashion MNIST \cite{xiao2017fashion}, EMNIST Letters \cite{cohen2017emnist}, and  CIFAR-10 \cite{krizhevsky2009learning}.
%These datasets consist of 12, 10, 26, and 10 classes, respectively.
%To train the model in each experiment, we sampled 10,000 data from the training split randomly per trial.
%This is to push the situation to an extreme; with 1/1000 or 1/500 query frequency restriction, the model can only get a maximum of 10 or 20 types of class labels, respectively.
%Therefore, the selection of representative samples becomes more crucial.
For experiments, we used four kinds of datasets with different distributions: Mouse retina transcriptomes \cite{macosko2015highly, polivcar2019opentsne}, Fashion MNIST \cite{xiao2017fashion}, EMNIST Letters \cite{cohen2017emnist}, and  CIFAR-10 \cite{krizhevsky2009learning}.
The datasets consist of 12, 10, 26, and 10 classes, respectively.
The budget $B$ was set to 1 or 2, and the period $W$ was set to 100, 500, 1,000 or 2,000.
For model training, we only used 10,000 randomly sampled data from the training split per trial.
This is to push the situation to an extreme; with 1/500 or less query frequency, the model will not be able to receive labels for some classes unless it successfully selects and queries all the samples from different classes.

\begin{figure*}[ht]
%\vskip 0.2in
\centering
    \subfigure[\footnotesize Mouse retina transcriptomes]{\includegraphics[width=0.24\textwidth]{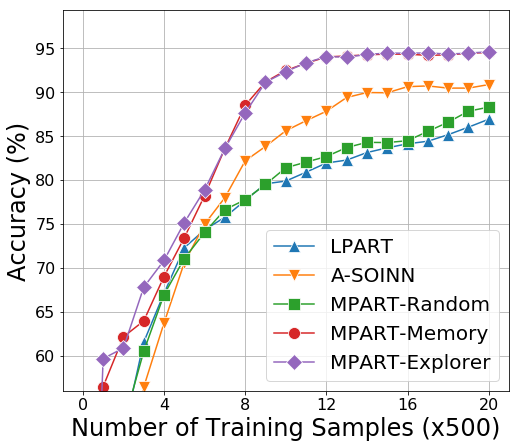}}
    \hfill
    \centering
    \subfigure[\footnotesize Fashion MNIST]{\includegraphics[width=0.24\textwidth]{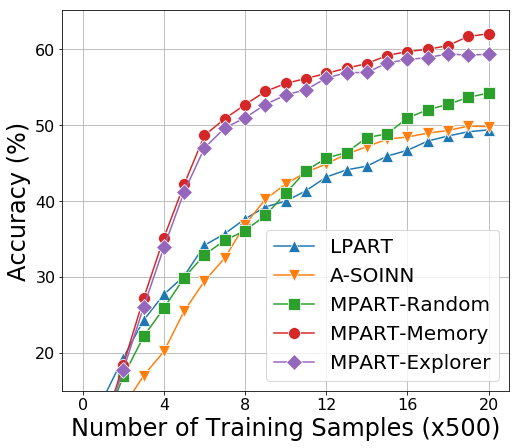}}
    \hfill
    \centering
    \subfigure[\footnotesize EMNIST Letters]{\includegraphics[width=0.24\textwidth]{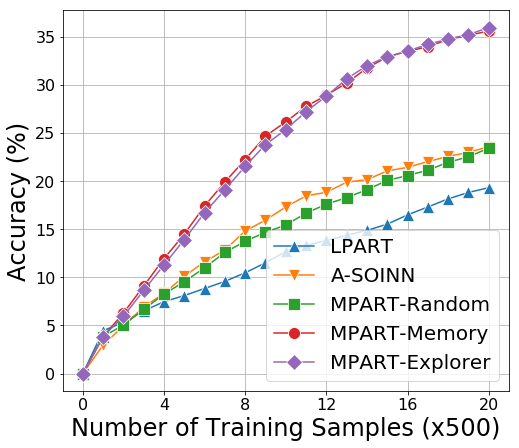}}
    \hfill
    \centering
    \subfigure[\footnotesize CIFAR-10]{\includegraphics[width=0.24\textwidth]{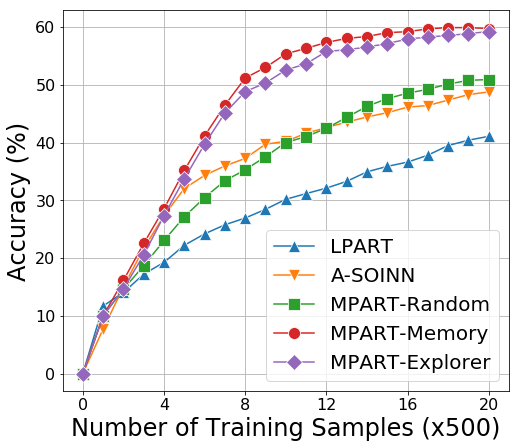}}
    \caption{Classification accuracy according to the number of training samples. The number of layer $L=3$ and 1 / 500 query frequency were used for MPART. The order of input samples was shuffled regardless of the class. Note that, for the EMNIST Letters dataset, the models can collect up to 20 class labels out of a total of 26 class labels after all training is complete.}
\label{fig:evolve}
%\vskip -0.2in
\end{figure*}

\subsection{Query Selection Strategy}
\label{sec:strategy}

\textbf{Random. }
A random query is selected from a sequence of inputs.
This is a baseline and works as an ablated version of active selection strategies.
%In this case, MPART operates in an online semi-supervised learning manner.
%In this case, MPART learns in an online semi-supervised manner.
It is equivalent to randomly providing labels for some inputs, so MPART learns in an online semi-supervised manner when using this strategy.

\textbf{Memory. }
%'Memory' is a strategy that can only be used at 1/$W$ query frequency, assuming the learning agent has a memory that can store only a single sample.
This can only be used at 1/$W$ query frequency, assuming the model has a memory that can store at most one sample.
%During each query period, one sample with the maximum  density-weighted query selection score $s_{t}$ is stored and inquired whenever the period ends.
During each query period, one sample with the maximum  density-weighted query selection score $s_{t}$ is stored and queried at the end of the period.
%If the label density of nodes is high enough, $u_{J,e}$ converges to 0, so $u_{J}$ changes according to $u_{J,a}$ and more data near the decision boundary is selected as a query target.
%By using this strategy, we can select the most beneficial data sample in the input stream.
By using this strategy, the model can select the most beneficial sample.
% 오라클이 가능할 때까지 쿼리할 샘플을 가지고 있다가, 쿼리 가능

\textbf{Explorer. }
`Explorer' assumes the most stringent situation where the learning agent cannot store any input sample.
In this situation, the learning agent selects $B$ samples online for each query period $W$, and cannot change the sample once selected.
Therefore, the chance of selecting an beneficial sample decreases as the exploration gets longer because of the fixed query selection period. 
In this strategy, the uncertainty distribution of the data explored so far is continuously estimated to solve the exploration-exploitation dilemma \cite{berger2014exploration}.
The random variable of the uncertainty distribution $\mathcal{S}_n$ is assumed to follow the normal distribution $\mathcal{S}_n \sim N(\mu_{n}, \sigma_{n}^2)$, where $n$ is the number of accumulated samples since the previous query. 
%$n$는 이전 query 이후 현재까지 accumulated된 sample들의 개수이다.
The parameters $\mu_{n}$ and $\sigma_{n}^2$ are calculated as the sample mean and variance of $s_t$ as in Equation \ref{eq:score_muvar}.
% using Equation \ref{eq:score_muvar}
%
\begin{equation} \label{eq:score_muvar}
\begin{split}
\mu_{n} & = (1 - n^{-1}) \mu_{n-1} + n^{-1}  s_{t} \\
\sigma_{n}^2 & = (1 - n^{-1})  \sigma_{n-1}^2 + n^{-1} (\mu_{n} - s_{t})^2 
\end{split}
\end{equation}
%

%Here, $s_{t}$ is the density-weighted query selection score of the input $x_t$.
%For an input sample, if its query selection score is highly likely to be greater than that of most unseen samples, we consider it to be informative.
An input sample is considered beneficial if its density-based query selection score is likely to be greater than that of most unseen samples.
To be more concrete, we select the $t$-th sample if Equation \ref{eq:explorer_criterion} is satisfied.
Here, $Binom$ denotes a probability mass function of a binomial distribution, $F_{\mathcal{S}_n}(\cdot)$ is the cumulative density function of $\mathcal{S}_n$, $t_p$ is the time from the start of the current period, and $b$ is the number of remaining query budget in the current period.
If $\sigma_{n}=0$, we set $F_{\mathcal{S}_n}(s_{t})$ to 0.5. 
\begin{equation} \label{eq:explorer_criterion}
%\begin{small}
\begin{gathered}
%\sum_{m=0}^{b-1} \binom{T-t_{p}}{m} (1-F_{\mathcal{S}_n}(s_{t}))^{m} F_{\mathcal{S}_n}(s_{t})^{(T-t_{p}-m)} > 0.5
\sum_{m=0}^{b-1} Binom \left(m ; W - t_{p}, 1-F_{\mathcal{S}_n}(s_{t}) \right) > 0.5
\end{gathered}
%\end{small}
\end{equation}
%
% $Binom(\cdot ; XXX, YYY)$ denotes a probability mass function of a binomial distribution $B(XXX, YYY)$.
%query가 발생하면 $\mathcal{S}_n$을 $\mu_{0} = 0,  \sigma_{0} = 0$으로 초기화한다.
When a query is selected, $b$ is reduced by 1, and both $\mu_{0}$ and $\sigma_{0}$ are reset to 0.
%\mathcal{S}_n$ is initialized to  = 0$.
%query budget $B$ 회의 query를 모두 마친 뒤에는 다음 period의 시작까지 query를 하지 않는다.
If $b$ is 0, no query is performed until the end of the period, and $b$ is reset to $B$ when a new period starts.
%$t_{end}$는 다음 쿼리 period의 종료까지의 잔여 timestep으로, 항상 $T\le t_{end} < 2T$를 만족한다.
%The value $T$ is the query period, and $t_{end}$ is the number of samples remaining until the end of the query period.%
% The agent selects the last sample if all samples are rejected until $t_{end}$.
%When the query period is ended, we reset $t$ for next period.%
With the `Explorer' strategy, the model can efficiently and effectively choose the useful samples by predicting the beneficialness of the unseen ones.
%오라클이 수행할 수 있는 

\subsection{Feature Extraction \& Dimension Reduction}
\label{sec:featext}
%Then, in order for MPART to properly form clusters and generate a topological graph for high-dimensional input data, we used pretrained dimension reducer Parametric UMAP \cite{sainburg2020parametric} for each dataset.
%Then we projected the input data into a 4-dimensional embedding space using the pre-trained Parametric UMAP \cite{sainburg2020parametric} for each dataset.
We projected the input data into 4-dimensional embedding space using the pre-trained Parametric UMAP \cite{sainburg2020parametric} for each dataset.
The embedding space was mapped to $[0,1]^4$ using min-max normalization on training data.
%First, we extracted 2048-dimensional feature vectors of the CIFAR-10 dataset using the BYOL \cite{grill2020bootstrap} model pre-trained with the ImageNet \cite{deng2009imagenet} dataset.
Prior to projection, we extracted 2048-dimensional feature vectors from CIFAR-10 using the BYOL \cite{grill2020bootstrap} pre-trained with the ImageNet dataset \cite{deng2009imagenet}.
%Parametric UMAP은 학습된 parametric mapping을 이용하여 fast online embeddings for new data, clustering에 적합하게 차원을 축소하여 줌으로써 semi-supervised learning에 적합하다.
Parametric UMAP is suitable for use in semi-supervised learning by capturing structure in unlabeled data to provide the embeddings for unseen data online which can be clustered properly.
%is useful for clustering.
%inputs using learned parametric mapping.
%이때 각 데이터셋의 훈련 split의 30\%를 이용하여 Parametric UMAP을 선학습하였다.
We trained Parametric UMAP using 30\% of the training data for each dataset, and the rest was used for MPART training.
Note that no labeled data was used until this step.
%In order for MPART to properly form clusters and generate a topological graph for high-dimensional input data, we used pre-trained feature extractor and dimension reducer.

\begin{table*}[t]
%\caption{The classification accuracy (mean $\pm$ std) of MPART depending on whether applying density-weighted query selection score (DS) and the message passing (MP) method is applied. The number of layer $L=3$ was used. (unit : \%)}
    \caption{Comparison of classification accuracy (mean $\pm$ std) between our model (MPART) and the competitive models according to the query selection strategy. We also report classification accuracy of MPART depending on whether density-weighted query selection score (DS) is applied or not. The number of layers $ L = 3 $ for message passing of MPART was used. (unit : \%)}
\label{tab:ablation_results}
%\vskip 0.15in
\begin{center}
\begin{small}
\begin{sc}
        %\begin{adjustbox}{max width=0.99\columnwidth}
        \begin{tabular}{c c c c c c}
        \toprule
        Query Selection & \multirow{2}{*}{Model} & Mouse retina & Fashion & EMNIST & \multirow{2}{*}{CIFAR-10} \\
        %\cmidrule(r){1-2}
        %\midrule
        Frequency & & transcriptomes & MNIST & Letters &  \\
        \midrule
\multirow{4}{*}{Fully Supervised} & MLP             & 94.0$\pm$1.7 & 71.9$\pm$1.6 & 50.9$\pm$2.0 & \textbf{75.1}$\pm$1.7 \\
                                  & LPART           & \textbf{97.2}$\pm$0.1 & 73.2$\pm$0.3 & 60.2$\pm$0.4 & 74.2$\pm$0.3 \\
                                  & A-SOINN         & 94.2$\pm$1.5 & 66.5$\pm$1.5 & 47.6$\pm$1.6 & 65.5$\pm$1.7 \\
                                  & MPART & \textbf{97.2}$\pm$0.1 & \textbf{73.3}$\pm$0.2 & \textbf{60.7}$\pm$0.3 & 74.3$\pm$0.3 \\
        \midrule
                          & LPART                      & 82.5$\pm$7.5 & 41.1$\pm$6.4 & 14.7$\pm$2.6 & 23.8$\pm$5.3 \\
                          & A-SOINN                    & 88.1$\pm$13.0 & 45.2$\pm$7.0 & 16.9$\pm$4.1 & 37.7$\pm$6.8 \\
                          \cmidrule(r){2-6}
\multirow{3}{*}{1 / 1000} & MPART-Random               & 81.0$\pm$7.0 & 42.8$\pm$8.5 & 17.1$\pm$3.1 & 40.9$\pm$8.1 \\
                          & MPART-Memory               & \textbf{93.3}$\pm$3.3 & \textbf{53.6}$\pm$4.8 & \textbf{26.1}$\pm$2.8 & \textbf{55.5}$\pm$4.6 \\
                          & MPART-Explorer             & 91.8$\pm$4.0 & 53.5$\pm$5.4 & 25.4$\pm$2.4 & 55.4$\pm$3.7 \\
                          \cmidrule(r){2-6}
                          %& MPART-Random (w/o MP)      & 30.7$\pm$6.1 & 13.0$\pm$0.9 & 4.3$\pm$0.2 & 10.7$\pm$0.3 \\
                          & MPART-Memory (w/o DS) & 90.5$\pm$5.6 & 51.4$\pm$5.6 & 19.8$\pm$3.5 & 44.9$\pm$7.0 \\
                          %& MPART-Memory (w/o MP) & 29.5$\pm$3.5 & 15.4$\pm$1.0 & 5.1$\pm$0.2 & 11.3$\pm$0.4 \\
                          %\cmidrule(r){2-6}
                          & MPART-Explorer (w/o DS)    & 87.2$\pm$8.1 & 49.9$\pm$6.9 & 18.2$\pm$3.4 & 44.1$\pm$6.7 \\
                          %& MPART-Explorer (w/o MP)    & 30.7$\pm$3.4 & 14.7$\pm$1.2 & 5.0$\pm$0.3 & 11.3$\pm$0.2 \\
\midrule
                          & LPART                      & 89.0$\pm$6.3 & 54.6$\pm$5.9 & 21.0$\pm$3.4 & 34.0$\pm$4.7 \\
                          & A-SOINN                    & 91.0$\pm$5.4 & 50.3$\pm$6.0 & 25.6$\pm$4.8 & 44.2$\pm$7.0 \\
                          \cmidrule(r){2-6}
\multirow{3}{*}{1 / 500}  & MPART-Random               & 88.3$\pm$4.4 & 54.3$\pm$7.6 & 23.4$\pm$3.8 & 50.9$\pm$5.2 \\
                          & MPART-Memory               & 94.5$\pm$1.1 & \textbf{62.0}$\pm$4.4 & 35.6$\pm$2.5 & \textbf{59.7}$\pm$3.4 \\
                          & MPART-Explorer             & \textbf{94.6}$\pm$0.9 & 59.4$\pm$5.2 & \textbf{36.0}$\pm$2.9 & 59.2$\pm$3.3 \\
                          \cmidrule(r){2-6}
                          %& MPART-Random (w/o MP)      & 28.5$\pm$3.9 & 16.0$\pm$1.4 & 4.9$\pm$0.3	& 11.5$\pm$0.4 \\
                          & MPART-Memory (w/o DS)      & 91.9$\pm$3.4 & 51.7$\pm$5.8 & 28.7$\pm$3.3 & 52.2$\pm$5.4 \\
                          %& MPART-Memory (w/o MP) & 31.1$\pm$1.6 & 19.7$\pm$1.6 & 6.1$\pm$0.3 & 12.7$\pm$0.4 \\
                          & MPART-Explorer (w/o DS)    & 92.1$\pm$3.4 & 53.9$\pm$4.8 & 29.2$\pm$3.8 & 51.6$\pm$6.6 \\
                          %& MPART-Explorer (w/o MP)    & 30.8$\pm$2.6 & 19.2$\pm$1.3 & 6.0$\pm$0.3 & 12.7$\pm$0.5 \\
\midrule
                          & LPART                      & 94.6$\pm$1.5 & 64.4$\pm$2.0 & 38.0$\pm$2.3 & 62.0$\pm$2.9 \\
                          & A-SOINN                    & 91.5$\pm$7.0 & 55.7$\pm$6.8 & 29.9$\pm$4.9 & 50.9$\pm$7.9 \\
                          \cmidrule(r){2-6}
\multirow{3}{*}{1 / 100}  & MPART-Random               & 94.8$\pm$1.2 & 66.8$\pm$2.0 & 43.6$\pm$3.2 & 63.4$\pm$2.6 \\
                          & MPART-Memory               & \textbf{95.9}$\pm$0.5 & \textbf{67.7}$\pm$1.6 & \textbf{47.9}$\pm$1.6 & \textbf{67.4}$\pm$1.5 \\
                          & MPART-Explorer             & \textbf{95.9}$\pm$0.9 & \textbf{67.7}$\pm$1.3 & 47.5$\pm$1.5 & 66.8$\pm$1.4 \\
                          \cmidrule(r){2-6}
                          %& MPART-Random (w/o MP)      & 44.9$\pm$2.7 & 33.5$\pm$2.5 & 8.8$\pm$0.5 & 16.5$\pm$0.7 \\
                          & MPART-Memory (w/o DS)      & 95.0$\pm$1.6 & 63.2$\pm$2.9 & 39.7$\pm$2.5 & 60.6$\pm$2.7 \\
                          %& MPART-Memory (w/o MP) & 57.5$\pm$1.8 & 43.6$\pm$1.7 & 12.5$\pm$0.5 & 20.8$\pm$0.6 \\
                          & MPART-Explorer (w/o DS) & 94.4$\pm$1.5 & 65.7$\pm$2.1 & 41.9$\pm$2.0 & 62.3$\pm$2.4 \\
                          %& MPART-Explorer (w/o MP) & 57.5$\pm$2.2 & 43.8$\pm$1.5 & 12.1$\pm$0.4 & 20.4$\pm$0.7 \\
        \bottomrule
        \end{tabular}
        %\end{adjustbox}
\end{sc}
\end{small}
\end{center}
%\vskip -0.1in
\end{table*}

\begin{table*}[t]
\caption{Classification accuracy (mean $\pm$ std) according to MPART's query frequency and number of layers $L$ for message passing. The `Explorer' strategy was used for query selection of MPART. (unit : \%)}
\label{tab:layer_results}
%\vskip 0.15in
\begin{center}
\begin{small}
\begin{sc}
        \begin{tabular}{c c c c c c c}
        \toprule
        Query Selection & Number of & Mouse retina & Fashion & EMNIST & \multirow{2}{*}{CIFAR-10} \\
        %\cmidrule(r){2-2}
        %\midrule
        Frequency & Layers & transcriptomes  & MNIST & Letters &  \\
        \midrule

\multirow{4}{*}{2 / 2000} %& LPART           & 79.5$\pm$8.0 & 39.7$\pm$7.3 & 13.7$\pm$2.7 & 30.7$\pm$5.2 \\
                          %& A-SOINN         & 90.6$\pm$9.0 & 46.8$\pm$5.0 & 18.7$\pm$3.2 & 40.8$\pm$7.2 \\
                          & $L = 5$ & 90.0$\pm$7.1 & 52.6$\pm$5.7 & 25.6$\pm$2.8 & 52.9$\pm$4.6 \\
                          & $L = 3$ & \textbf{92.3}$\pm$4.0 & \textbf{56.5}$\pm$3.9 & \textbf{26.5}$\pm$2.3 & \textbf{53.9}$\pm$4.4 \\
                          & $L = 1$ & 66.9$\pm$4.7 & 44.0$\pm$5.2 & 14.3$\pm$1.6 & 27.3$\pm$2.2 \\
                          & w/o MP  & 31.1$\pm$4.9 & 15.2$\pm$0.9 & 4.9$\pm$0.3 & 11.3$\pm$0.3 \\
         \midrule
\multirow{4}{*}{4 / 2000} %& LPART           & 89.1$\pm$4.8 & 50.4$\pm$6.4 & 20.5$\pm$3.2 & 42.8$\pm$6.3 \\
                          %& A-SOINN        & 91.7$\pm$8.0 & 54.1$\pm$4.4 & 26.7$\pm$4.5 & 51.6$\pm$6.7 \\
                          & $L = 5$ & \textbf{94.1}$\pm$2.2 & \textbf{60.1}$\pm$4.6 & 33.4$\pm$3.2 & 58.1$\pm$4.0 \\
                          & $L = 3$ & 93.1$\pm$3.5 & 60.0$\pm$4.4 & \textbf{36.0}$\pm$2.8 & \textbf{59.5}$\pm$3.8 \\
                         & $L = 1$ & 86.8$\pm$2.6 & 59.2$\pm$3.9 & 21.4$\pm$1.2 & 38.3$\pm$2.0 \\
                         & w/o MP & 31.0$\pm$2.3 & 19.6$\pm$1.6 & 6.0$\pm$0.2 & 12.5$\pm$0.5 \\
                        \midrule
\multirow{4}{*}{20 / 2000} %& LPART           & 94.3$\pm$1.7 & 64.5$\pm$2.1 & 39.0$\pm$2.5 & 60.1$\pm$2.4 \\
                         %& A-SOINN       & 93.7$\pm$2.3 & 64.2$\pm$2.6 & 40.2$\pm$3.1 & 61.2$\pm$4.3 \\
                         & $L = 5$ & \textbf{96.1}$\pm$0.5 & \textbf{67.6}$\pm$1.5 & 47.6$\pm$1.6 & \textbf{67.4}$\pm$1.6 \\
                         & $L = 3$ & 95.8$\pm$0.7 & 67.3$\pm$1.5 & \textbf{48.0}$\pm$1.7 & 67.2$\pm$1.3 \\
                         & $L = 1$ & 95.0$\pm$1.1 & 66.6$\pm$1.4 & 40.3$\pm$1.5 & 59.1$\pm$1.1 \\
                         & w/o MP & 57.8$\pm$1.5 & 43.6$\pm$2.2 & 12.2$\pm$0.6 & 20.5$\pm$0.6 \\
        \bottomrule
        \end{tabular}
\end{sc}
\end{small}
\end{center}
%\vskip -0.1in
\end{table*}

\section{Results and Discussion}

%In this case, the input data was projected into a 2-dimensional embedding space for visualization.

\subsection{Comparison with Competitive Models}

%The performance of the competitive models such as A-SOINN \cite{shen2011incremental} and LPART \cite{kim2020label} was compared to MPART using various message passing layers $L$ and `Explorer' strategy in Table \ref{tab:layer_results}.
The performance of the competitive models LPART \cite{kim2020label} and A-SOINN \cite{shen2011incremental} was compared to that of MPART with various query selection strategy, shown in Table \ref{tab:ablation_results} and Figure \ref{fig:evolve}.
%The multi-layer perceptron (MLP) model was used as a reference of fully supervised learning, which was trained using all labeled data for each dataset.
The fully supervised settings including multi-layer perceptron (MLP) model were used as references, which was trained using all labeled data for each dataset.
%For comparison, Fuzzy ART related parameters in LPART used the same values as MPART, and parameters of A-SOINN were adjusted to show the highest performance.
The MLP is consisted of 3 layers with 128 neurons per layer and we reported the highest test accuracy while training up to 200 epochs.
The same ART-related parameter values were used in MPART and LPART, while A-SOINN parameters were adjusted to achieve its best performance for fair comparison.
A-SOINN needs to query the prototype of the most dense node, which is a weighted sum of encoded representations and does not correspond to any input samples.
Therefore, in A-SOINN, the prototype of the node was not directly queried, but the most recent input sample that activates the node was queried.
In all experimental settings except with MLP, the MPART showed the highest accuracy.
%, and even outperformed the MLP on the Mouse retina transcriptomes dataset.
%It was confirmed that the class distribution of the Mouse retina transcriptomes dataset is imbalanced, and the number of training data was insufficient, so it showed low performance in the MLP.

%%%%%
\textbf{Comparison with LPART. }
MPART with `Random' strategy showed higher performance than LPART, which performs online semi-supervised learning.
This indicates that although both models are based on the same backbone model, Fuzzy ART, MPART's message passing method is more effective than LPART's label propagation method.
%This indicates that the message passing method of MPART is more effective than LPART, which does not fully utilize the topology information of data.
MPART's massage passing differs from LPART's label propagation in several aspects.
The critical difference is that LPART does not sufficiently learn and use topology information of input data.
%Moreover, in LPART, information is exchanged between all co-activated nodes, whereas in MPART, edge connection is strengthened only between the winning node and the rest of the co-activated nodes.
%This allows the amount of information transmitted between two nodes to be appropriately adjusted according to the edge connection strength $e_{ij}$, regardless of the connection strength with other nodes.
LPART permanently updates the label density $q$ when label propagation is performed, whereas MPART updates $q^0$ using only true labels and infers the class through message passing.
The MPART's message passing method with multi-layers prevents information propagation of one node from permanently affecting other nodes, and when a new label is acquired, the information can be immediately reflected in the inference of all nodes.
In addition, MPART queries representative samples by estimating the distribution density of input data using the topology learned online, whereas in the case of LPART, it is difficult to select a representative sample because it cannot effectively estimate the distribution of the input data.
%이런 이유로 fig.5 (c) ~(d) 에서 보면 준지도 지속학습을 수행하는 MPART (Random) 이 LPART 보다 더 좋은 성능을 보임

\textbf{Comparison with A-SOINN. }
A-SOINN employs an explicit removal of edges and nodes to reduce the noise and maintain a reasonable model size.
The removal procedure heavily depends on the deletion period $\lambda$, which should ideally be determined according to the input statistics.
Since such information is not known \textit{a priori}, a wrong choice of $\lambda$ severely degrades the model performance.
%We could actually see a performance trade-off at different query budgets when changing $\lambda$.
Also, each node in A-SOINN holds only one label which is fixed since the first assignment either with actual label or by propagation from other nodes.
Therefore, the order of labeling highly influences the inference results, especially at decision boundaries.
This might explain why A-SOINN even performs worse than LPART in fully-supervised setting.

%\vskip 0.1in
\textbf{Computational  Cost. }
MPART requires more computation than LPART and A-SOINN, but the computational cost required to update the model is very low.
We measured the run-time of our Python implementation on a 3.8 GHz CPU machine.
The time taken to train and infer 10,000 of CIFAR-10 data samples was 11.2 seconds on average, which took about 1.12 ms per sample.
%Also, since we only store the representative values in a graph, the number of nodes and edges created is significantly less than the number of training data.
For the CIFAR-10 dataset, the average numbers of nodes are 1260, 1922, and 2442 when trained with 15k, 30k and 45k samples respectively.
%The storage space of the MPART model trained with 10,000 of CIFAR-10 data samples is 3.8 MB on average, which is much lower than the 29.3 MB storage space of the trained raw data.
%In this case, the storage space of the MPART is 3.8 MB on average, which is much lower than the 29.3 MB storage space of the trained raw data.
%In addition, the proposed method considers not only the storage space but also the environments in which data storage is prohibited due to privacy issues.
Please refer to the Appendix for detailed statistical analysis and additional experimental results.

\subsection{Ablation Study}

Table \ref{tab:ablation_results} summarizes the results of performance evaluation of our model on four datasets according to query selection frequencies and strategies.
In the w/o DS setting, the query selection score $u_t$ of Equation \ref{eq:u_sum} was used instead of $s_t$ of Equation \ref{eq:score}.
%In all experimental settings, the model that applies density-weighted query selection score (DS) and message passing (MP) showed significantly higher performance than the one without.
In all experimental settings, the model that applies density-weighted query selection score showed significantly higher performance than the one without (w/o DS).
%The accuracies decrease as the frequency of query decreases, but the gap is not significant when the message passing method is applied.
When comparing the performance with respect to the query selection strategy, the accuracy of the `Random' strategy was generally low, and the `Memory' strategy and the `Explorer' strategy showed almost similar performance.
This is because the `Explorer' strategy properly estimates the uncertainty distribution of input data and efficiently selects the representative samples based on the remaining query opportunities.
%In addition, the lower the query frequency, the greater the performance difference according to the query selection strategy, indicating that the query selection strategy can have a significant impact on performance in a situation where the labeled data is extremely scarce.
In addition, the lower the query frequency, the greater the difference in performance depending on the query selection strategy, indicating that the strategy can have a significant impact on performance in situations where labeled data is extremely scarce.
Table \ref{tab:layer_results} shows the performance evaluation of MPART according to the number of layers $L$ for message passing.
As the number of layers increased, the classification performance generally increased, but the results of using 3 and 5 layers were almost the same.
On the other hand, when message passing is not applied (w/o MP), there is a significant performance penalty.

\section{Conclusions}

%We propose Message Passing Adaptive Resonance Theory (MPART) that learns the distribution and the topology of the input data online, infers the class of unlabeled data, and selects the informative and representative samples through message passing between nodes on the topological graph.
We propose Message Passing Adaptive Resonance Theory (MPART) for online active semi-supervised learning.
MPART learns the distribution and the topology of the input data online, infers the class of unlabeled data, and selects the informative and representative samples through message passing between nodes on the topological graph.
By evaluating our method on datasets including EMNIST Letters and CIFAR-10, we show that it outperforms the competitive models.
In an online learning environment where data storage is limited and data labeling is expensive, waste of useful samples is inevitable.
The proposed model fully utilizes the underlying structure of input data so that it can minimize the waste.
This approach also reduces the need to create large datasets in advance in order to apply machine learning to various industries.
We believe MPART offers new opportunities for machine learning techniques to be widely used in real-world applications.

%\begin{comment}
\section*{Acknowledgements}

This work was partly supported by the Institute of Information \& Communications Technology Planning \& Evaluation (2019-0-01371-BabyMind/35\%, 2015-0-00310-SW.StarLab/5\%, 2017-0-01772-VTT/5\%, 2018-0-00622-RMI/5\%) grant funded by the Korean government.
%This work was partly supported by the Korea government (2019-0-01367-BabyMind, 2015-0-00310-SW.StarLab, 2017-0-01772-VTT, 2018-0-00622-RMI, P0006720-GENKO).
%\end{comment}

%\bibliography{refs}
\bibliography{refs}
\bibliographystyle{icml2021}

\onecolumn
\renewcommand\thesection{\Alph{section}}
\setcounter{section}{0}

\section{Appendix}
\label{appendix}

\subsection{Dataset}

\begin{figure}[ht!]
\vskip 0.2in
\centering
\subfigure[\footnotesize Mouse retina transcriptomes]{\includegraphics[width=0.48\textwidth]{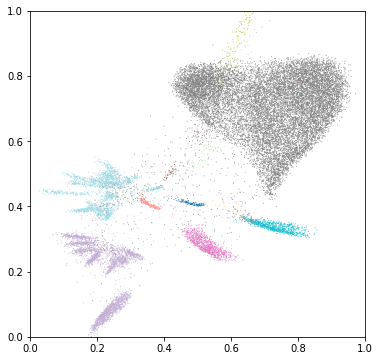}}
\hfill
\centering
\subfigure[\footnotesize Fashion MNIST]{\includegraphics[width=0.48\textwidth]{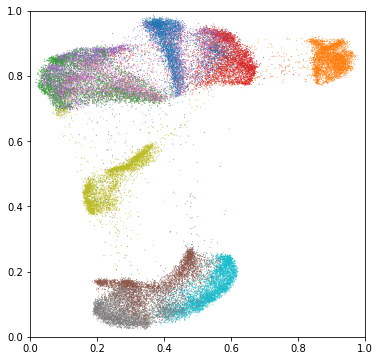}}
\hfill

\subfigure[\footnotesize EMNIST Letters]{\includegraphics[width=0.48\textwidth]{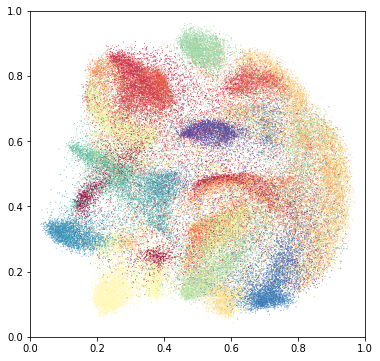}}
\hfill
\centering
\subfigure[\footnotesize CIFAR-10]{\includegraphics[width=0.48\textwidth]{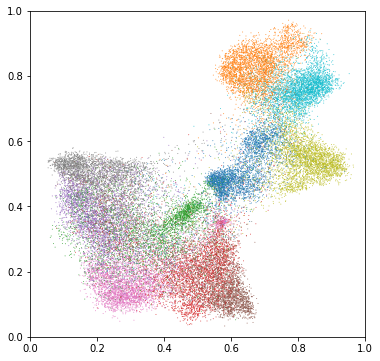}}
\caption{Visualization of the distribution of each dataset using Parametric UMAP.}

\label{fig:distribution}
\vskip -0.2in
\end{figure}

We used 4 types of datasets: Mouse retina transcriptomes \cite{macosko2015highly, polivcar2019opentsne}, Fashion MNIST \cite{xiao2017fashion}, EMNIST Letters \cite{cohen2017emnist}, and  CIFAR-10 \cite{krizhevsky2009learning}. The details on the datasets and the numbers used in training and evaluations are shown in Table \ref{tab:dataset}.
%The MNIST dataset consists of handwritten digit images from 0 to 9.
%The SVHN is a real-world image dataset obtained from house numbers.
%The NSynth is an audio dataset containing musical notes, each with a unique pitch, timbre, and envelope.
The Mouse retina transcriptomes dataset consists of PCA projections of single-cell transcriptome data collected from mouse retina.
The Fashion MNIST dataset consists of 10 types of clothing such as shirts, sneakers, and ankle boots.
The EMNIST Letters dataset is a set of handwritten alphabet characters of 26 classes.
The CIFAR-10 dataset consists of real-world images in 10 classes such as airplanes, automobiles, birds and cats.
Except for Mouse retina transcriptomes dataset, the distribution of classes is uniform.
The class distribution of the Mouse retina transcriptomes dataset is shown in Table \ref{tab:mouse_dataset}.
The distribution of each dataset is visualized in Figure \ref{fig:distribution}.
For visualization, each dataset was projected into a 2-dimensional embedding space using Parametric UMAP.

At least 10,000 samples of each dataset were used as test dataset for the performance evaluation of MPART.
We also used 30\% of the training dataset to train the Parametric UMAP \cite{sainburg2020parametric} for dimensionality reduction.
MPART was trained using only 10,000 randomly sampled data from the remaining 70\% of the training dataset.

\begin{table*}[h!]
\caption{Datasets used to evaluate our proposed method and the number of samples used in training and testing.}
\label{tab:dataset}
    \vskip 0.15in
    \begin{center}
    \begin{small}
    \begin{sc}
        \begin{tabular}{c c c c c c}
        \toprule
        \multirow{2}{*}{Dataset} & \multirow{2}{*}{Num. of Classes} & \multirow{2}{*}{Dimensions} & Parametric UMAP & \multicolumn{2}{c}{MPART} \\
        \cmidrule(r){5-6}
         &  &  & (Train) & Train & Test \\
        \midrule
        Mouse retina transcriptomes & 12 & 50 & 10,442 & 24,366 & 10,000\\
        Fashion MNIST & 10 & 28x28 & 18,000 & 42,000 & 10,000 \\
        EMNIST Letters & 26 & 28x28 & 37,440 & 87,360 & 20,800 \\
        CIFAR-10 & 10 & 32x32x3 & 15,000 & 35,000 & 10,000 \\
        \bottomrule
        \end{tabular}
        \end{sc}
    \end{small}
    \end{center}
    \vskip -0.1in
\end{table*}

\begin{table*}[h!]
\caption{Distribution of classes in Mouse retina transcriptomes dataset.}
\label{tab:mouse_dataset}
    \vskip 0.15in
    \begin{center}
    \begin{small}
    \begin{sc}
        \begin{tabular}{c c c c c c}
        \toprule
        \multirow{2}{*}{Label} & \multirow{2}{*}{Class Name} & Parametric UMAP & \multicolumn{2}{c}{MPART} & \multirow{2}{*}{Total} \\
        \cmidrule(r){4-5}
        & & (Train) & Train & Test &  \\
        \midrule
        0 & Cones & 425 & 1,029 & 414 & 1,868 \\
        1 & Horizontal cells & 67 & 131 & 54 & 252 \\
        2 & Pericytes & 16 & 37 & 10 & 63 \\
        3 & Amacrine cells & 1,041 & 2,314 & 1,071 & 4,426 \\
        4 & Retinal ganglion cells & 113 & 228 & 91 & 432 \\
        5 & Fibroblasts & 21 & 50 & 14 & 85 \\
        6 & Vascular endothelium & 54 & 139 & 59 & 252 \\
        7 & Muller glia & 363 & 893 & 368 & 1,624\\
        8 & Astrocytes & 15 & 29 & 10 & 54 \\
        9 & Microglia & 15 & 41 & 11 & 67 \\
        10 & Rods & 6,855 & 16,047 & 6,498 & 29,400 \\
        11 & Bipolar cells & 1,457 & 3,428 & 1,400 & 6,285 \\
        \midrule
        & Total & 10,442 & 24,366 & 10,000 & 44,808 \\
        \bottomrule
        \end{tabular}
        \end{sc}
\end{small}
\end{center}
\vskip -0.1in
    
\end{table*}

The datasets can be downloaded from the following links.
\begin{itemize}
\item Mouse retina transcriptomes: \url{http://file.biolab.si/opentsne/macosko_2015.pkl.gz}
\item Fashion MNIST: \url{https://github.com/zalandoresearch/fashion-mnist}
\item EMNIST Letters: \url{https://github.com/aurelienduarte/emnist}
\item CIFAR-10: \url{https://www.cs.toronto.edu/~kriz/cifar.html}
\end{itemize}

%\textbf{Visualization of Class Distribution. }
\newpage

\newpage

\subsection{Implementation Details}
\label{sec:implementation_details}

\textbf{Parametric UMAP. }
%The limitation of MPART is that it cannot directly handle high-dimensional input data, and the performance of the pre-trained dimensionality reduction method can affect the overall performance.
%We used state-of-the-art techniques for dimensionality reduction because the performance of pre-trained dimensionality reduction methods can affect overall performance.
In order for MPART to cluster input data and properly generate the topological graph, it is necessary to represent high-dimensional input data in a low-dimensional latent space.
So, the performance of the dimensionality reduction can affect the overall performance.
We extracted the 2048-dimensional feature vector of the CIFAR-10 dataset using the BYOL \cite{grill2020bootstrap} model pretrained with the ImageNet \cite{deng2009imagenet} dataset.
Then we projected the input data into a 4-dimensional embedding space using the pretrained Parametric UMAP for each dataset.
The Parametric UMAP was trained using the parameters shown in Table \ref{tab:umap_parameter}, and the training was conducted for 50 epochs.
A convolutional neural network (CNN) was used as an embedding algorithm of Parametric UMAP for Fashion MNIST and EMNIST Letters.
For the Mouse retina transcriptome and CIFAR-10, a 3-layer multi-layer perceptron (MLP) was used with 100-neurons per layer.
More information about Parametric UMAP can be found at the following link: \url{https://umap-learn.readthedocs.io/}.

\begin{table}[h!]
\caption{Summary of important UMAP parameters used in this study.}
\label{tab:umap_parameter}
    \vskip 0.15in
    \begin{center}
    \begin{small}
    \begin{sc}
        \begin{tabular}{c c c}
        \toprule
        Name & Value & Description \\
        %\cmidrule(r){-9}
        % & Train & Test & Train & Test & Train & Test & Train & Test \\
        \midrule
        n\_neighbors & 15 & controls how UMAP balances local versus global structure in the data \\
        \midrule
        min\_dist & 0.1 & controls how tightly UMAP is allowed to group points together  \\
        \midrule
        n\_components & 4 & controls the dimensionality of the reduced dimension space  \\
        \midrule
        metric & euclidean & the method for distance measure in the ambient space of the input data  \\
        \bottomrule
        \end{tabular}
    \end{sc}
    \end{small}
    \end{center}
    %\vskip -0.1in
\end{table}

%n_neighbors=15, min_dist=0.1, n_components=2, metric='euclidean'

\textbf{MPART. }
We chose the parameters of the MPART empirically.
The parameters used in the model are shown in Table \ref{tab:parameters}, and the same values were used for all datasets and experiments.
The value of $\rho$ should be determined by considering the number of dimensions of the input space and computational power.
The larger $\rho$ value, the more MPART nodes are created, which allows for finer clustering, but increases the required computational power.
The value of $k_e$ should be set according to how many queries are possible.
A small value of $k_e$ should be used in situations where enough queries are possible.
Since we dealt with very few labeled data, we used a relatively large value of $k_e = 1.0$.
The value of $k_d$ should be set small enough according to the total amount of training data including unlabeled data.
We also show in Table \ref{tab:parameters_others} the parameters used in the competitive models.
% In LPART, ART related parameters have the same values as MPART.

%To train the MPART in each experiment, we randomly sampled 10,000 data from the training set for each trial.
%All data samples were inputted only once, one by one.

%We set the vigilance parameter $\rho$ to $0.97$ for MNIST and Nsynth datasets, and $0.95$ for SVHN and CIFAR-10 datasets.
%Unless specified otherwise, we used these values for all experiments.
%$\rho$ means the intolerance for getting new data, so we used higher $\rho$ for clearer datasets to get better performance. We used $\alpha=0.01$ (choice parameter), $\beta=0.5$ (learning rate of ART), $\delta=0.5$ (propagation rate), $k=0.2$ (sensitivity parameter), $\tau=0.3$ (weight for query selection score) for all experiments.

%We pretrained feature extractor with learning rate 0.001. We trained ??? epochs for each dataset. Feature extractor was optimized by Adam optimizer.

\begin{table}[h!]
    \caption{Model parameters used in the MPART.}
    \label{tab:parameters}
    \vskip 0.15in
    \begin{center}
    \begin{small}
    \begin{sc}
        \begin{tabular}{c c c}
        \toprule
        Symbol & Value & Description \\
        %\cmidrule(r){-9}
        % & Train & Test & Train & Test & Train & Test & Train & Test \\
        \midrule
        $ \alpha $ & 0.01 & choice parameter  \\
        \midrule
        $ \beta $ & 0.5 & learning rate for node weights \\
        \midrule
        $ \rho $ & 0.95 & vigilance parameter \\
        \midrule
        $ \delta $ & 0.1 & propagation rate for message passing \\
        \midrule
        $ \tau $ & 0.7 & weight for query selection score \\
        \midrule
        $ k_e $ & 1.0 & sensitivity for epistemic uncertainty \\
        \midrule
        $ k_d $ & 0.01 & sensitivity for density-weighted query selection score \\
        \midrule
        $ L $ & 3 & number of layers for message passing \\
        \bottomrule
        \end{tabular}
    \end{sc}
    \end{small}
    \end{center}
    \vskip -0.1in
\end{table}

\begin{table}[h!]
    \caption{Model parameters used in LPART and A-SOINN.}
    \label{tab:parameters_others}
    \vskip 0.15in
    \begin{center}
    \begin{small}
    \begin{sc}
        \begin{tabular}{c c c c}
        \toprule
        
        Model & Symbol & Value & Description\\
        \midrule % LPART
         & $ \alpha $ & 0.01 & choice parameter  \\
        \cmidrule(r){2-4}
        \multirow{2}{*}{LPART} & $ \beta $ & 0.5 & learning rate for node weights \\
        \cmidrule(r){2-4}
        & $ \rho $ & 0.95 & vigilance parameter \\
        \cmidrule(r){2-4}
        & $ \delta $ & 1.0 & propagation rate \\
        \midrule % A-SOINN
        & $ \lambda $ & 500 & period for node removal and clustering \\
        \cmidrule(r){2-4}
        \multirow{1}{*}{A-SOINN} & $ \mathrm{age}_{\mathrm{max}} $ & 30 & maximum age of edge \\
        \cmidrule(r){2-4}
        & $ \alpha $ & 2.0 & smoothing parameter for grouping \\
        \bottomrule
        \end{tabular}
    \end{sc}
    \end{small}
    \end{center}
    \vskip -0.1in
\end{table}

%\vskip 0.3in
\newpage
\subsection{Parameter Search for $\delta$ and $\tau$}
The choice of values of the propagation rate $\delta$ in message passing and the weight $\tau$ for query selection score affects the performance only marginally, unless $\delta$ is set to 0.
Table \ref{tab:para_result} shows the classification performance on CIFAR-10 according to the change of $\delta$ and $\tau$.
The `Explorer' strategy and 1/500 query frequency were used.
The mean and standard deviation were drawn from results of 10 trials.

\begin{table*}[h!]
\caption{The classification accuracy (mean $\pm$ std) of our model with various $\delta$ and $\tau$. For each $L$, \textbf{boldface} indicates the top three and the \textcolor{blue}{blue text} indicates the bottom three.}
    \label{tab:para_result}
    \vskip 0.15in
    \begin{center}
    \begin{small}
    \begin{sc}
        \begin{tabular}{c c c c c c c c c}
        \toprule
        Number of & \multirow{3}{*}{$\delta$} & \multicolumn{7}{c}{$\tau$} \\
        \cmidrule(r){3-9}
        layers &  & 0.0 & 0.1 & 0.3 & 0.5 & 0.7 & 0.9 & 1.0 \\
        \midrule
        $L=0$                  & 0.0 & 12.5$\pm$0.4 & 12.6$\pm$0.3 & 12.8$\pm$0.3 & 12.7$\pm$0.4 & 12.6$\pm$0.5 & 12.8$\pm$0.6 & 12.8$\pm$0.4 \\
        \midrule
        \multirow{6}{*}{$L=1$} & 0.1 & 39.2$\pm$1.9 & 38.6$\pm$1.8&39.5$\pm$1.9&39.3$\pm$2.0&39.3$\pm$1.9&\textcolor{blue}{35.8}$\pm$1.7&\textcolor{blue}{33.6}$\pm$2.6 \\
                               & 0.3 & 39.2$\pm$1.9 & 38.7$\pm$2.2&39.7$\pm$2.2&39.2$\pm$1.8&38.4$\pm$2.4&38.4$\pm$1.9&\textcolor{blue}{34.2}$\pm$1.6 \\
                               & 0.5 & 39.3$\pm$1.4 & 39.5$\pm$2.3&39.3$\pm$1.6&39.1$\pm$1.4&38.4$\pm$1.7&38.3$\pm$2.2&37.2$\pm$1.5 \\
                               & 0.7 & 38.2$\pm$2.4 & 38.7$\pm$1.7&\textbf{39.8}$\pm$3.1&37.8$\pm$2.3&\textbf{39.7}$\pm$1.4&37.8$\pm$1.8&38.7$\pm$2.8 \\
                               & 0.9 & 39.3$\pm$2.4 & 38.1$\pm$2.5&38.8$\pm$2.6&37.0$\pm$2.9&37.2$\pm$2.2&38.8$\pm$2.0&38.4$\pm$1.6 \\
                               & 1.0 & 39.1$\pm$3.3 & 38.4$\pm$2.4&37.1$\pm$3.3&\textbf{40.3}$\pm$2.0&38.1$\pm$2.0&36.7$\pm$3.9&38.2$\pm$2.3 \\
        \midrule
        \multirow{6}{*}{$L=3$} & 0.1 & 61.4$\pm$3.8 & 61.1$\pm$1.7 & \textbf{63.4}$\pm$2.4 & 59.3$\pm$3.6 & 58.9$\pm$2.9 & 56.7$\pm$3.4 & \textcolor{blue}{55.2}$\pm$3.5 \\
                               & 0.3 & 59.5$\pm$5.8 & 58.8$\pm$3.8 & \textbf{61.7}$\pm$3.3 & 59.3$\pm$3.9 & 59.3$\pm$3.5 & 58.5$\pm$5.9 & 59.3$\pm$3.0 \\
                               & 0.5 & 57.9$\pm$5.5 & 58.8$\pm$4.0 & 59.8$\pm$4.9 & 60.5$\pm$3.4 & 60.5$\pm$2.8 & 59.2$\pm$3.6 & 57.5$\pm$4.6 \\
                               & 0.7 & 60.4$\pm$2.5 & \textcolor{blue}{56.3}$\pm$5.2 & 60.4$\pm$3.7 & 59.9$\pm$3.1 & 58.7$\pm$3.1 & 59.9$\pm$3.8 & 60.0$\pm$4.2 \\
                               & 0.9 & 58.0$\pm$4.8 & \textcolor{blue}{56.4}$\pm$5.3 & 58.4$\pm$5.9 & 58.9$\pm$3.1 & 60.0$\pm$3.8 & 60.0$\pm$3.0 & \textbf{61.5}$\pm$3.9 \\
                               & 1.0 & 58.9$\pm$4.3 & 58.6$\pm$4.5 & 59.2$\pm$5.2 & 56.6$\pm$5.9 & 61.0$\pm$3.5 & 59.7$\pm$3.9 & 59.5$\pm$3.7 \\
        \midrule
        \multirow{6}{*}{$L=5$} & 0.1 & 54.6$\pm$6.6 & 55.1$\pm$7.8 & 60.1$\pm$2.7 & 60.1$\pm$2.6 & 60.0$\pm$4.0 & 57.4$\pm$2.3 & 60.0$\pm$3.4 \\
                               & 0.3 & 50.6$\pm$4.4 & 51.2$\pm$6.6 & 54.3$\pm$3.8 & \textbf{61.5}$\pm$2.1 & \textbf{60.5}$\pm$4.5 & 60.3$\pm$4.3 & \textbf{60.8}$\pm$3.7 \\
                               & 0.5 & 49.7$\pm$7.3 & 53.8$\pm$4.6 & 56.1$\pm$5.2 & 59.1$\pm$3.7 & 58.9$\pm$3.5 & 59.1$\pm$4.9 & 58.9$\pm$3.6 \\
                               & 0.7 & \textcolor{blue}{48.2}$\pm$4.9 & 52.0$\pm$2.9 & 53.8$\pm$8.0 & 57.5$\pm$5.3 & 58.2$\pm$3.8 & 60.4$\pm$4.4 & 60.5$\pm$2.2 \\
                               & 0.9 & \textcolor{blue}{47.2}$\pm$6.3 & \textcolor{blue}{45.7}$\pm$8.6 & 52.1$\pm$4.4 & 58.0$\pm$4.1 & 56.3$\pm$5.7 & 57.2$\pm$6.3 & 56.8$\pm$5.8 \\
                               & 1.0 & 49.1$\pm$4.3 & 51.0$\pm$4.0 & 51.3$\pm$4.2 & 53.5$\pm$7.9 & 56.2$\pm$10.4 & 54.0$\pm$7.4 & 58.9$\pm$2.1 \\
        \bottomrule
        \end{tabular}
    \end{sc}
    \end{small}
    \end{center}
    %\vskip -0.1in
    
\end{table*}

\newpage
\subsection{Additional Experimental Results}

\begin{figure}[ht!]
\vskip 0.05in
\begin{center}
\centerline{\includegraphics[width=0.99\columnwidth]{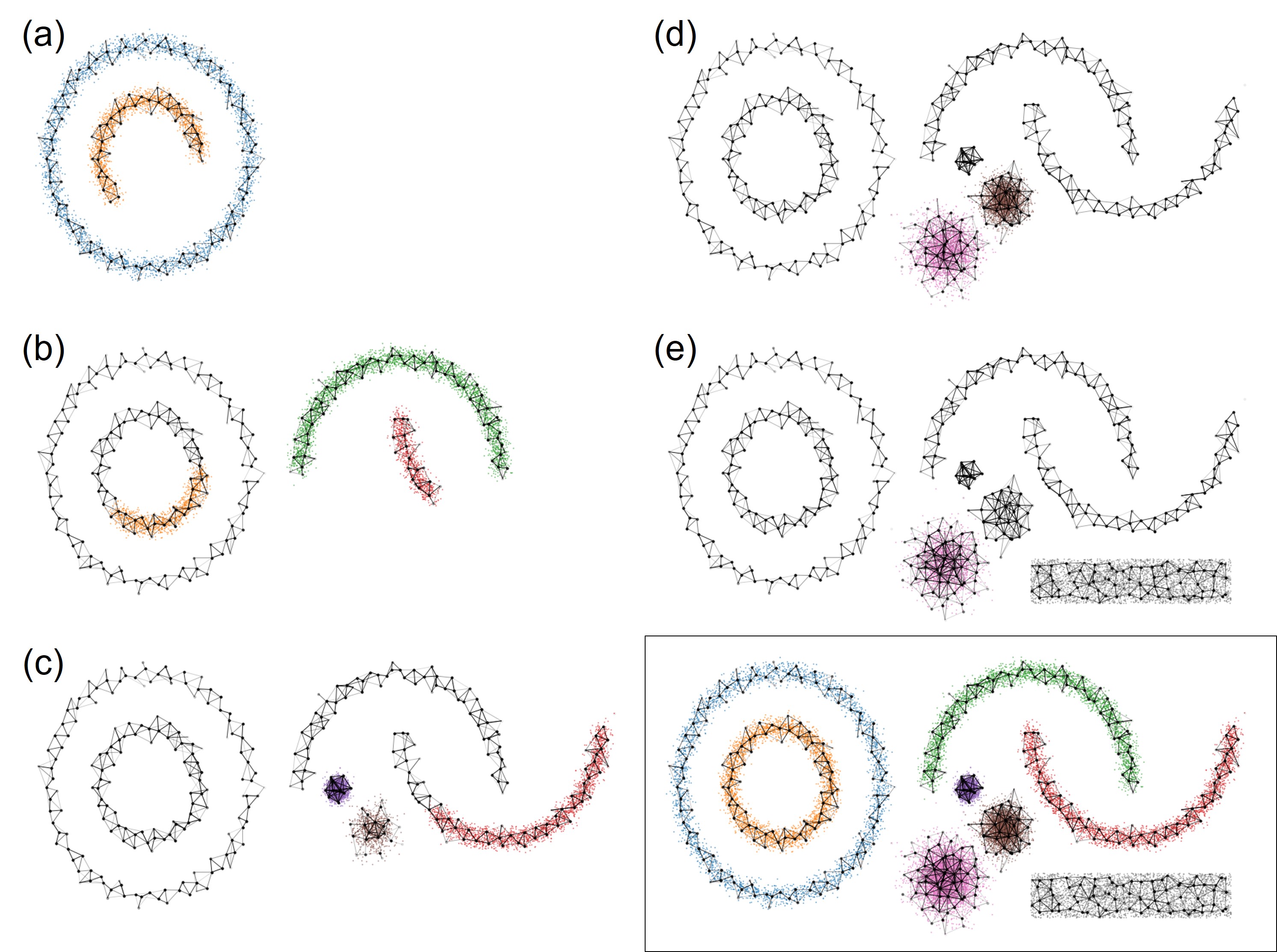}}
\caption{An example result of topology learning. (a) $\sim$ (e) is the process of topology learning, and the lower right figure is the final learning result. The colored scattered points refer to newly entered data and the intensity of the topological graph represents the density of nodes and edges. All data samples were input only once, one by one.}
\label{fig:topology}
\end{center}
\vskip -0.2in
\end{figure}

\textbf{Online Topology Learning. }
Figure \ref{fig:topology} shows an example result of MPART's online topology learning.
MPART continuously learns the distribution and topology structure of sequential input data without forgetting.

\newpage
\textbf{Analysis of the Generated Topological Graph. }
For a deeper analysis, it would be interesting to see some statistics of the topological graph.
In Table \ref{tab:statistic}, we show the numbers of generated nodes, co-activations for each task, and co-activated nodes per winning node and per sample.
We also analyzed the average number of neighboring nodes per node, weights of edges, and the nodes that do not have any co-activated node, in which case will be indicated by nodes without edges.

\begin{table*}[h!]
\caption{Detailed analysis of the trained topological graph. The mean and standard deviation are drawn from 10 trials for each dataset.}
\label{tab:statistic}
    \vskip 0.15in
    \begin{center}
    \begin{small}
    \begin{sc}
        \begin{tabular}{c c c c c}
        \toprule
        \multirow{2}{*}{Item} & Mouse retina & \multirow{2}{*}{Fashion MNIST} & EMNIST & \multirow{2}{*}{CIFAR-10} \\
         & transcriptomes &  & Letters &  \\
        \midrule
        Total \# of Nodes                      & 290.3$\pm$5.3     & 256.2$\pm$4.3     & 1165.5$\pm$8.0   & 984.6$\pm$10.5 \\
        Total \# of Co-activations             & 15873.2$\pm$399.5 & 15533.1$\pm$404.4 & 9222.8$\pm$151.4 & 11230.6$\pm$210.6  \\
        Avg. \# of Co-activations per Node     & 54.69$\pm$1.68    & 60.63$\pm$1.48    & 7.91$\pm$0.15    & 11.41$\pm$0.24 \\
        Avg. \# of Co-activations per Sample   & 1.59$\pm$0.04     & 1.55$\pm$0.04     & 0.92$\pm$0.02    & 1.12$\pm$0.02 \\
        Avg.  \# of Neighboring Nodes per Node & 7.46$\pm$0.12     & 7.51$\pm$0.22     & 4.16$\pm$0.06    & 6.02$\pm$0.08 \\
        Total \# of Nodes w/o Edges            & 11.00$\pm$3.43    & 6.70$\pm$2.11     & 34.20$\pm$6.07   & 19.50$\pm$3.03 \\
        Avg. \# of Weights of Edges            & 0.179$\pm$0.009   & 0.193$\pm$0.007   & 0.172$\pm$0.003  & 0.150$\pm$0.003 \\
        \bottomrule
        \end{tabular}
        \end{sc}
    \end{small}
    \end{center}
    \vskip -0.1in
\end{table*}

\vskip 0.2in

\textbf{Computational Cost. }
We measured the run-time of our Python implementation on a 3.8 GHz CPU machine.
Tables \ref{tab:node} and \ref{tab:time} show the number of nodes generated by MPART and the time required for training and inference according to the number of training samples.
%As the amount of training data increases, the number of generated nodes increases, but the time required for training and inference per sample increases only slightly.

\begin{table*}[h!]
\caption{The number of nodes generated by MPART according to the query selection frequency and the number of training samples. The number of layer $L = 3$ and the `Explorer' strategy were used. The mean and standard deviation are drawn from 10 trials for each dataset.}
\label{tab:node}
\vskip 0.1in
\begin{center}
\begin{small}
\begin{sc}
        \begin{tabular}{c c c c c c c}
        \toprule
        Query selection & Number of & Mouse retina & Fashion & EMNIST & \multirow{2}{*}{CIFAR-10} \\
        %\cmidrule(r){2-2}
        %\midrule
        frequency & training data & transcriptomes  & MNIST & Letters &  \\
        \midrule
        \multirow{3}{*}{1 / 1000} & 10,000 & 290.2$\pm$5.0 & 254.8$\pm$4.3 & 1170.6$\pm$9.4 & 970.7$\pm$7.4 \\
                                  & 15,000 & 349.0$\pm$5.8 & 301.0$\pm$4.1 & 1504.4$\pm$9.6 & 1260.6$\pm$6.7 \\
                                  & 20,000 & 396.1$\pm$4.4 & 338.1$\pm$4.8 & 1810.3$\pm$6.9 & 1505.7$\pm$7.8 \\
        \midrule
        \multirow{3}{*}{1 / 500} & 10,000 & 290.5$\pm$4.5 & 254.3$\pm$5.3 & 1167.1$\pm$8.9 & 980.8$\pm$5.9 \\
                                  & 15,000 & 348.6$\pm$6.0 & 301.6$\pm$3.7 & 1511.6$\pm$11.3 & 1262.1$\pm$3.8 \\
                                  & 20,000 & 394.9$\pm$5.7 & 337.7$\pm$5.3 & 1810.6$\pm$8.5 & 1510.7$\pm$7.3 \\
        \midrule
        \multirow{3}{*}{1 / 100} & 10,000 & 290.5$\pm$4.6 & 255.0$\pm$4.5 & 1163.5$\pm$8.2 & 976.4$\pm$8.5 \\
                                  & 15,000 & 347.3$\pm$4.6 & 301.3$\pm$5.7 & 1510.7$\pm$10.8 & 1262.7$\pm$9.0 \\
                                  & 20,000 & 398.1$\pm$5.7 & 337.8$\pm$4.2 & 1807.6$\pm$8.4 & 1508.6$\pm$7.6 \\
        \bottomrule
        \end{tabular}
\end{sc}
\end{small}
\end{center}
\vskip -0.1in
\end{table*}

\begin{table*}[h!]
\caption{Comparison of time taken by MPART for training and inference according to query selection frequency and the number of training samples. The number of layer $L = 3$ and the `Explorer' strategy were used. The mean and standard deviation are drawn from 10 trials for each dataset. (unit: sec)}
\label{tab:time}
\vskip 0.1in
\begin{center}
\begin{small}
\begin{sc}
        \begin{tabular}{c c c c c c c}
        \toprule
        Query selection & Number of & Mouse retina & Fashion & EMNIST & \multirow{2}{*}{CIFAR-10} \\
        %\cmidrule(r){2-2}
        %\midrule
        frequency & training data & transcriptomes  & MNIST & Letters &  \\
        \midrule
        \multirow{3}{*}{1 / 1000} & 10,000 & 9.24$\pm$0.14 & 9.03$\pm$0.35 & 13.25$\pm$0.58 & 11.21$\pm$0.15 \\
                                  & 15,000 & 14.46$\pm$0.14 & 13.98$\pm$0.11 & 25.54$\pm$0.17 & 20.70$\pm$0.23 \\
                                  & 20,000 & 19.49$\pm$0.13 & 19.12$\pm$0.11 & 43.71$\pm$0.33 & 34.38$\pm$0.26 \\
        \midrule
        \multirow{3}{*}{1 / 500} & 10,000 & 9.08$\pm$0.10 & 8.90$\pm$0.07 & 13.03$\pm$0.11 & 11.43$\pm$0.20 \\
                                  & 15,000 & 14.55$\pm$0.11 & 13.93$\pm$0.05 & 25.84$\pm$0.32 & 20.90$\pm$0.21 \\
                                  & 20,000 & 19.83$\pm$0.08 & 19.12$\pm$0.17 & 43.60$\pm$0.41 & 34.59$\pm$0.19 \\
        \midrule
        \multirow{3}{*}{1 / 100} & 10,000 & 9.15$\pm$0.09 & 8.87$\pm$0.06 & 13.26$\pm$0.17 & 11.55$\pm$0.17 \\
                                  & 15,000 & 13.93$\pm$0.05 & 13.68$\pm$0.07 & 25.93$\pm$0.26 & 21.11$\pm$0.16 \\
                                  & 20,000 & 18.78$\pm$0.06 & 18.71$\pm$0.10 & 43.65$\pm$0.39 & 34.72$\pm$0.26 \\
        \bottomrule
        \end{tabular}
\end{sc}
\end{small}
\end{center}
\vskip -0.1in
\end{table*}

\newpage
\textbf{Various Query Frequencies. }
%Table \ref{tab:frequency} summarizes the results of performance evaluation of our model on four datasets according to query selection frequencies and strategies.
Table \ref{tab:frequency} summarizes the performance evaluation of MPART and competitive models on four datasets according to various query selection frequencies.
Given the same total query budget, MPART shows similar performance regardless of query frequencies.

\begin{table*}[h!]
\caption{Comparison of classification accuracy (mean $\pm$ std) between MPART and the competitive models. The number of layer $L = 3$ and the `Explorer' strategy were used for MPART. The mean and standard deviation are drawn from 30 trials for each dataset. (unit : \%)}
\label{tab:frequency}
\vskip 0.1in
\begin{center}
\begin{small}
\begin{sc}
        \begin{tabular}{c c c c c c c c}
        \toprule
        Total Query & Query Selection& \multirow{2}{*}{Model} & Mouse retina & Fashion & EMNIST & \multirow{2}{*}{CIFAR-10} \\
        %\cmidrule(r){2-2}
        %\midrule
        Budget & Frequency &  & transcriptomes  & MNIST & Letters &  \\
        \midrule
 & \multirow{3}{*}{1 / 1000}                     & LPART   & 77.1$\pm$7.4 & 41.5$\pm$4.4 & 13.3$\pm$3.2 & 27.4$\pm$5.0 \\
                    &                            & A-SOINN & 88.1$\pm$13.0 & 45.2$\pm$7.0 & 16.9$\pm$4.1 & 37.7$\pm$6.8 \\
                    &                            & MPART   & \textbf{91.2}$\pm$5.6 & \textbf{56.4}$\pm$5.8 & \textbf{26.4}$\pm$2.7 & \textbf{52.6}$\pm$4.9 \\
                    \cmidrule(r){2-7}
                    & \multirow{3}{*}{2 / 2000} & LPART   & 83.8$\pm$6.0 & 37.6$\pm$4.9 & 14.8$\pm$3.1 & 28.0$\pm$3.4 \\
                    &                            & A-SOINN & 90.9$\pm$5.0 & 45.4$\pm$7.9 & 17.8$\pm$2.7 & 42.7$\pm$7.0 \\
\multirow{2}{*}{10} &                            & MPART   & \textbf{91.7}$\pm$4.0 & \textbf{56.8}$\pm$3.7 &            \textbf{25.8}$\pm$1.9 & \textbf{54.8}$\pm$4.0 \\
                    \cmidrule(r){2-7}
                    & \multirow{3}{*}{5 / 5000} & LPART   & 84.3$\pm$9.9 & 41.6$\pm$7.4 & 15.4$\pm$2.8 & 30.1$\pm$3.7 \\
                    &                            & A-SOINN & \textbf{93.6}$\pm$2.6 & 49.8$\pm$5.2 & 21.5$\pm$2.5 & 46.2$\pm$6.2 \\
                    &                            & MPART   & 92.2$\pm$4.4 & \textbf{55.0}$\pm$3.7 & \textbf{26.2}$\pm$1.7 & \textbf{54.4}$\pm$5.3 \\
                    \cmidrule(r){2-7}
                    & \multirow{3}{*}{10 / 10000} & LPART   & 81.3$\pm$4.2 & 38.4$\pm$6.4 & 15.6$\pm$3.7 & 36.0$\pm$6.5 \\
                    &                            & A-SOINN & \textbf{94.4}$\pm$2.0 & 50.3$\pm$4.7 & 23.2$\pm$2.6 & 50.8$\pm$4.6 \\
                    &                            & MPART   & 91.8$\pm$4.7 & \textbf{54.6}$\pm$4.6 & \textbf{25.4}$\pm$2.1 & \textbf{56.0}$\pm$4.3 \\
                    \midrule
                    
                    & \multirow{3}{*}{1 / 500} & LPART   & 85.4$\pm$4.3 & 49.3$\pm$7.4 & 21.2$\pm$2.7 & 42.0$\pm$5.7 \\
                    &                            & A-SOINN & 91.0$\pm$5.4 & 50.3$\pm$6.0 & 25.6$\pm$4.8 & 44.2$\pm$7.0 \\
                    &                            & MPART   & \textbf{93.5}$\pm$2.6 & \textbf{61.4}$\pm$3.1 & \textbf{35.9}$\pm$3.4 & \textbf{59.2}$\pm$2.6 \\
                    \cmidrule(r){2-7}
                    & \multirow{3}{*}{2 / 1000} & LPART   & 89.4$\pm$4.8 & 52.3$\pm$6.3 & 20.7$\pm$2.9 & 42.1$\pm$7.9 \\
                    &                            & A-SOINN & 92.0$\pm$3.7 & 51.9$\pm$4.6 & 26.0$\pm$4.0 & 49.3$\pm$6.1 \\
                    &                            & MPART   & \textbf{94.3}$\pm$1.8 & \textbf{59.8}$\pm$4.1 & \textbf{35.7}$\pm$3.4 & \textbf{59.1}$\pm$3.1 \\
                    \cmidrule(r){2-7}
                    & \multirow{3}{*}{4 / 2000} & LPART   & 86.5$\pm$5.5 & 54.8$\pm$4.0 & 22.9$\pm$3.6 & 42.8$\pm$6.9 \\
                 20 &                            & A-SOINN & 93.3$\pm$3.2 & 55.0$\pm$5.1 & 28.4$\pm$3.9 & 51.9$\pm$4.9 \\
                    &                            & MPART   & \textbf{93.5}$\pm$2.9 & \textbf{60.8}$\pm$4.6 & \textbf{35.7}$\pm$2.8 & \textbf{60.0}$\pm$3.7 \\
                    \cmidrule(r){2-7}
                    & \multirow{3}{*}{10 / 5000} & LPART   & 89.5$\pm$6.9 & 50.1$\pm$7.0 & 19.3$\pm$3.0 & 41.7$\pm$6.2 \\
                    &                            & A-SOINN & \textbf{94.2}$\pm$2.3 & 56.4$\pm$4.7 & 31.9$\pm$2.8 & 57.8$\pm$3.8 \\
                    &                            & MPART   & 93.9$\pm$1.8 & \textbf{59.7}$\pm$4.5 & \textbf{36.7}$\pm$2.1 & \textbf{60.4}$\pm$3.2 \\
                    \cmidrule(r){2-7}
                    & \multirow{3}{*}{20 / 10000} & LPART   & 89.8$\pm$4.5 & 49.4$\pm$4.4 & 23.0$\pm$3.0 & 45.4$\pm$6.8 \\
                    &                            & A-SOINN & \textbf{94.6}$\pm$1.8 & 59.8$\pm$4.0 & 36.2$\pm$3.1 & \textbf{60.9}$\pm$4.0 \\
                    &                            & MPART   & 93.4$\pm$2.5 & \textbf{60.1}$\pm$3.8 & \textbf{36.9}$\pm$4.3 & 59.9$\pm$4.0 \\
                    \midrule
                    
                    & \multirow{3}{*}{1 / 100} & LPART   & 94.4$\pm$0.9 & 65.8$\pm$1.8 & 37.7$\pm$2.3 & 58.2$\pm$1.9 \\
                    &                            & A-SOINN & 91.5$\pm$7.0 & 55.7$\pm$6.8 & 29.9$\pm$4.9 & 50.9$\pm$7.9 \\
                    &                            & MPART   & \textbf{95.7}$\pm$0.8 & \textbf{67.3}$\pm$1.7 & \textbf{47.9}$\pm$1.8 & \textbf{67.0}$\pm$1.4 \\
                    \cmidrule(r){2-7}
                    & \multirow{3}{*}{10 / 1000} & LPART   & 94.5$\pm$1.3 & 64.4$\pm$1.8 & 37.8$\pm$2.4 & 59.8$\pm$2.5 \\
                    &                            & A-SOINN & 93.9$\pm$1.6 & 63.0$\pm$3.8 & 39.5$\pm$3.0 & 60.3$\pm$5.8 \\
                    &                            & MPART   & \textbf{95.6}$\pm$0.9 & \textbf{67.5}$\pm$1.4 & \textbf{47.7}$\pm$1.7 & \textbf{67.4}$\pm$1.1 \\
                    \cmidrule(r){2-7}
                    & \multirow{3}{*}{20 / 2000} & LPART   & 93.1$\pm$1.4 & 64.7$\pm$2.2 & 38.8$\pm$2.4 & 60.0$\pm$2.0 \\
                100 &                            & A-SOINN & 93.6$\pm$2.3 & 63.9$\pm$3.0 & 40.6$\pm$3.8 & 60.6$\pm$4.3 \\
                    &                            & MPART   & \textbf{95.7}$\pm$0.9 & \textbf{66.7}$\pm$1.9 & \textbf{47.6}$\pm$1.3 & \textbf{66.8}$\pm$1.5 \\
                    \cmidrule(r){2-7}
                    & \multirow{3}{*}{50 / 5000} & LPART   & 93.9$\pm$1.6 & 63.4$\pm$2.4 & 37.2$\pm$2.6 & 59.4$\pm$1.7 \\
                    &                            & A-SOINN & 93.6$\pm$2.4 & 64.3$\pm$2.8 & 42.2$\pm$3.3 & 60.5$\pm$3.9 \\
                    &                            & MPART   & \textbf{95.4}$\pm$1.1 & \textbf{67.3}$\pm$1.5 & \textbf{47.8}$\pm$1.9 & \textbf{66.7}$\pm$1.5 \\
                    \cmidrule(r){2-7}
                    & \multirow{3}{*}{100 / 10000} & LPART   & 94.9$\pm$1.1 & 64.1$\pm$2.3 & 37.7$\pm$1.8 & 58.4$\pm$2.8 \\
                    &                            & A-SOINN & 94.2$\pm$1.3 & 64.8$\pm$3.1 & 41.9$\pm$2.8 & 63.6$\pm$4.1 \\
                    &                            & MPART   & \textbf{95.5}$\pm$0.7 & \textbf{67.6}$\pm$1.2 & \textbf{47.3}$\pm$1.3 & \textbf{66.7}$\pm$1.3 \\
        \bottomrule
        \end{tabular}
\end{sc}
\end{small}
\end{center}
\vskip -0.1in
\end{table*}

\begin{figure}[h!]
\vskip 0.2in
\centering
\subfigure[\footnotesize Random]{\includegraphics[width=0.33\textwidth]{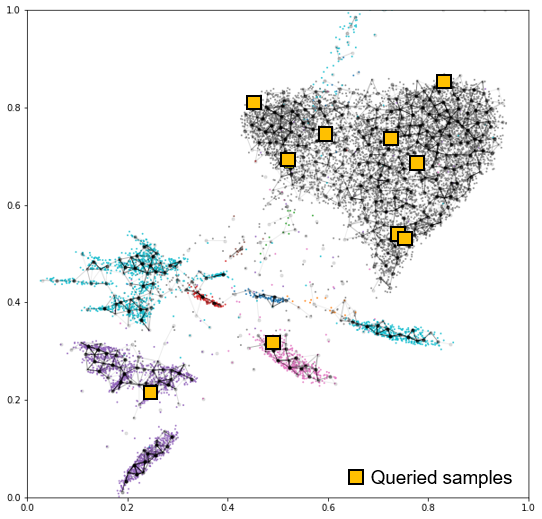}}
\hfill
\centering
\subfigure[\footnotesize Memory]{\includegraphics[width=0.33\textwidth]{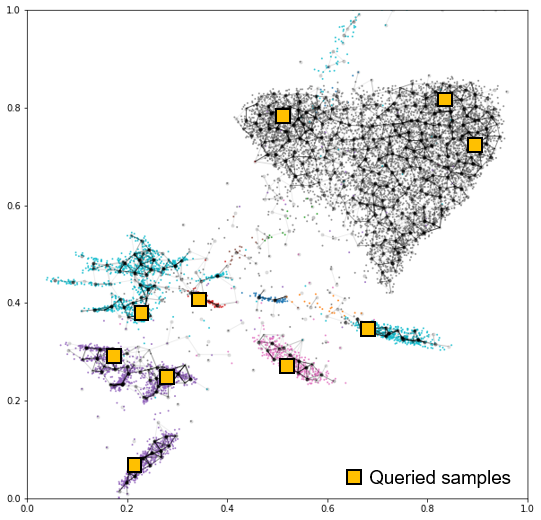}}
\hfill
\centering
\subfigure[\footnotesize Explorer]{\includegraphics[width=0.33\textwidth]{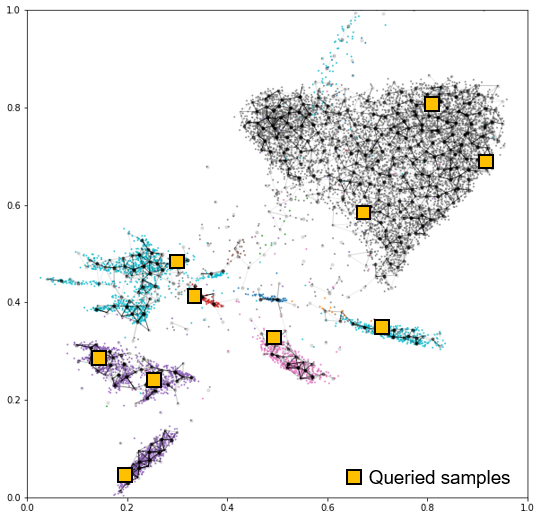}}
\caption{The visualization of training results on the Mouse retina transcriptomes dataset according to query selection strategy. The number of layer $ L= 3 $ and 1/1000 query frequency were used. The colored scattered points refer to input samples and the intensity of the topological graph represents the density of nodes and edges. The queried samples are indicated in yellow boxes. In the case of Random strategy, we can observe more queried samples in the class represented by the black dots. In contrast, the queried samples are more evenly distributed for each class when using the `Memory' or `Explorer' strategy. All data samples were input only once, one for each time step. The input data was projected into a 2-dimensional embedding space for visualization.}
\label{fig:macosko_strategy}
\vskip -0.2in
\end{figure}

\begin{figure}[h!]
\vskip 0.4in
\centering
\subfigure[\footnotesize Random]{\includegraphics[width=0.33\textwidth]{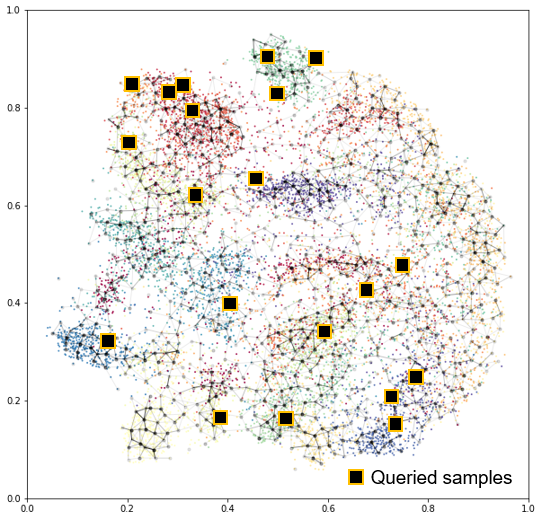}}
\hfill
\centering
\subfigure[\footnotesize Memory]{\includegraphics[width=0.33\textwidth]{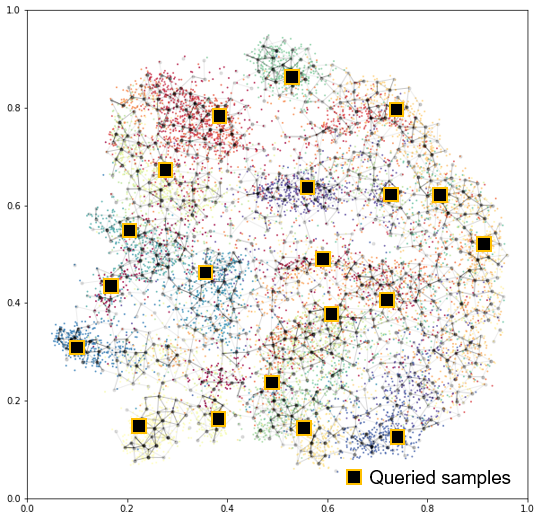}}
\hfill
\centering
\subfigure[\footnotesize Explorer]{\includegraphics[width=0.33\textwidth]{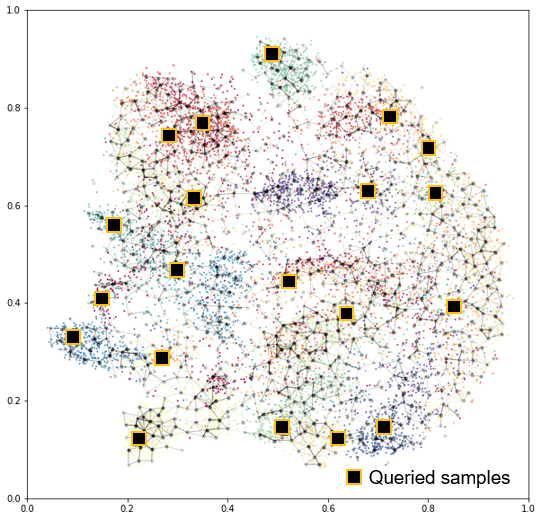}}
\caption{The visualization of training results on the EMNIST Letters dataset according to query selection strategy. The number of layer $ L= 3 $ and 1/500 query frequency were used. The colored scattered points refer to input samples and the intensity of the topological graph represents the density of nodes and edges. The queried samples are indicated in black boxes. In the case of Random strategy, we can observe more queried samples in certain classes than others. In contrast, the queried samples are more evenly distributed for each class when using the `Memory' or `Explorer' strategy. All data samples were input only once, one for each time step. The input data was projected into a 2-dimensional embedding space for visualization.}
\label{fig:emnist_strategy}
\vskip -0.2in
\end{figure}

\newpage
\paragraph{Visualization of Training Results. }
In order to evaluate the capability of our model for online active semi-supervised learning, we visualized the generated topological graph and query distributions.
The visualization results on the Mouse retina transcriptomes and EMNIST Letters datasets are shown in Figures \ref{fig:macosko_strategy} and \ref{fig:emnist_strategy}.
The visualization shows that the Memory and Explorer strategies perform better than the Random strategy, as implied by the evenly distributed queried samples.

%\newpage

%\bibliography{refs_appendix}
%\bibliographystyle{icml2021}

\end{document}